\documentclass[Afour,sageh,times]{sagej}

\usepackage{moreverb,url}

\usepackage[colorlinks,bookmarksopen,bookmarksnumbered,citecolor=blue,urlcolor=blue,linkcolor=blue ]{hyperref}

\newcommand\BibTeX{{\rmfamily B\kern-.05em \textsc{i\kern-.025em b}\kern-.08em
T\kern-.1667em\lower.7ex\hbox{E}\kern-.125emX}}

\def\volumeyear{2025}

\usepackage[nameinlink]{cleveref}
\usepackage{bm}
\usepackage{amsmath}
\usepackage{mathtools}
\usepackage{siunitx}
\usepackage[table]{xcolor}
\usepackage{tikz}
\usetikzlibrary{shapes, arrows, positioning, fit, calc}
\usepackage{placeins}
\usepackage{subcaption}

\newcommand{\M}[0]{\bm{M}}
\newcommand{\C}[0]{\bm{C}}
\newcommand{\K}[0]{\bm{K}}
\newcommand{\f}[0]{\bm{f}}

\newcommand{\fint}[0]{\bm{f}_{\mathrm{int}}}

\newcommand{\fact}[0]{\bm{f}_{\mathrm{act}}}

\newcommand{\udot}[0]{\dot{\bm{u}}}
\newcommand{\udd}[0]{\ddot{\bm{u}}}
\newcommand{\x}[0]{\bm{x}}
\newcommand{\xo}[0]{\bm{x}_0}

\newcommand{\E}[0]{\bm{E}}
\newcommand{\D}[0]{\bm{D}}
\newcommand{\Dt}[0]{\bm{D}^\top}

\newcommand{\Dxi}[0]{\bm{D}_\xi}
\newcommand{\Dxit}[0]{\bm{D}_\xi^\top}

\newcommand{\F}[0]{\bm{F}}
\newcommand{\I}[0]{\bm{I}}

\newcommand{\disp}[0]{\bm{u}}
\newcommand{\ud}[0]{\bm{u}_d}
\newcommand{\uxi}[0]{\bm{u}_{\xi}}

\newcommand{\V}[0]{\bm{V}}
\newcommand{\U}[0]{\bm{U}}

\newcommand{\etav}[0]{\bm{\eta}}
\newcommand{\etad}[0]{\dot{\bm{\eta}}}
\newcommand{\etadd}[0]{\ddot{\bm{\eta}}}
\newcommand{\etadnv}[0]{\dot{\eta}}
\newcommand{\etaddnv}[0]{\ddot{\eta}}
\newcommand{\xiv}[0]{\bm{\xi}}

\newcommand{\thetav}[0]{\bm{\theta}}

\newcommand{\phiv}[0]{\bm{\phi}}

\newcommand{\R}[0]{\bm{R}}

 \newcommand{\sens}[0]{\bm{S}}
\newcommand{\fd}[0]{\bm{f}_{\mathrm{drag}}}

\newcommand{\ft}[0]{\bm{f}_{\mathrm{thrust}}}
\newcommand{\ftail}[0]{\bm{f}_{\mathrm{tail}}}
\newcommand{\fspine}[0]{\bm{f}_{\mathrm{spine}}}

\newcommand{\nv}[0]{\bm{n}}
\newcommand{\T}[0]{\bm{T}}
\newcommand{\ze}[0]{\bm{0}}
\newcommand{\mt}[0]{\tilde{m}}
\newcommand{\A}[0]{\bm{A}}
\newcommand{\B}[0]{\bm{B}}

\newcommand{\pop}[2]{\frac{\partial{#1}}{\partial{#2}}}

\newcommand{\der}[2]{\dfrac{\mathrm{d}#1}{\mathrm{d}#2}}

\newcommand{\ps}[2]{\prescript{#1}{#2}}
\newcommand{\G}[0]{\bm{G}}
\newcommand{\m}[0]{\bm{m}}
\newcommand{\Hm}[0]{\bm{H}}
\newcommand{\Lm}[0]{\bm{L}}

\newcommand{\dvec}[0]{\bm{d}}

\definecolor{color1}{rgb}{0, 0.4470, 0.7410}
\definecolor{color2}{rgb}{0.8500 0.3250 0.0980}
\definecolor{color3}{rgb}{0.9290 0.6940 0.1250}

\newcommand{\SOOneSymbol}{
  \tikz[baseline=-0.7ex] \draw[fill=color1] (0,0) circle (3.7pt)
    node[cross out, draw=black, line width=0.4pt, minimum size=5pt, inner sep=0pt] {};
}
\newcommand{\SOTwoSymbol}{
    \tikz[baseline=-0.7ex] \draw[fill=color2] (0,0) circle (3.7pt)
    node[cross out, draw=black, rotate=45, line width=0.4pt, minimum size=5pt, inner sep=0pt] {};
}

\newcommand{\SOThreeSymbol}{
    \tikz[baseline=-0.7ex] \draw[fill=color3] (0,0) circle (3.7pt) ;
}

\def\MD{\textcolor{black}}

\begin{document}

\runninghead{Dubied et al.}

\title{AquaROM: shape optimization pipeline for soft swimmers using parametric reduced order models}

\author{Mathieu Dubied\affilnum{1,2}, Paolo Tiso\affilnum{2}, Robert K. Katzschmann\affilnum{1}}

\affiliation{\affilnum{1}Soft Robotics Laboratory (SRL), ETH Zurich, Switzerland\\
\affilnum{2}Chair in Nonlinear Dynamics, ETH Zurich, Switzerland
}

\corrauth{Robert K. Katzschmann, SRL, ETH Zurich, Switzerland}

\email{rkk@ethz.ch}

\begin{abstract}
The efficient optimization of actuated soft structures, particularly under complex nonlinear forces, remains a critical challenge in advancing robotics. Simulations of nonlinear structures, such as soft-bodied robots modeled using the finite element method (FEM), often demand substantial computational resources, especially during optimization. To address this challenge, we propose a novel optimization algorithm based on a tensorial parametric reduced order model (PROM). Our algorithm leverages dimensionality reduction and solution approximation techniques to facilitate efficient solving of nonlinear constrained optimization problems. The well-structured tensorial approach enables the use of analytical gradients within a specifically chosen reduced order basis (ROB), significantly enhancing computational efficiency. To showcase the performance of our method, we apply it to optimizing soft robotic swimmer shapes. These actuated soft robots experience hydrodynamic forces, subjecting them to both internal and external nonlinear forces, which are incorporated into our optimization process using a data-free ROB for fast and accurate computations. This approach not only reduces computational complexity but also unlocks new opportunities to optimize complex nonlinear systems in soft robotics, paving the way for more efficient design and control.
\end{abstract}

\keywords{Modeling, Control, and Learning for Soft Robots, Soft Robot Materials and Design, Optimization and Optimal Control}

\maketitle

\section{Introduction}

When designing mechanical systems, such as robots, MEMS, or wind turbines, engineers typically aim to find structures that exhibit optimal properties while respecting specific constraints. Instead of solely relying on testing physical prototypes of a given system, numerical simulations of these same prototypes are often less costly, faster, and easier to set up. In addition, simulations can provide useful information, such as gradients, to perform optimization of the system's properties in an automated fashion.

\begin{figure}[htb]
    \centering
    \includegraphics[width=0.94\linewidth]{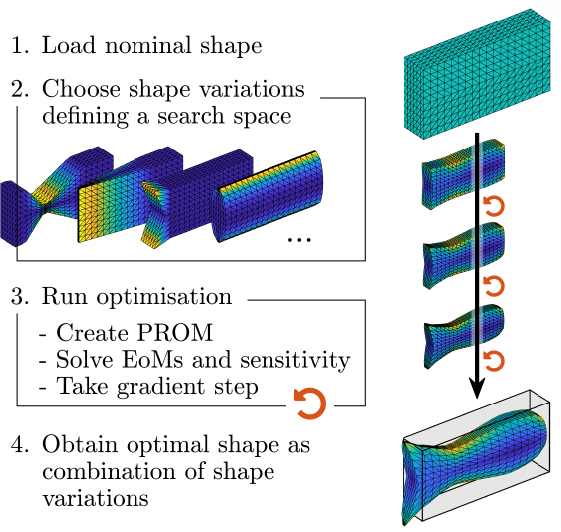}
    \caption{Optimization pipeline applied to the case of soft robotic fishes. Starting from a nominal shape, our pipeline solves a constrained optimization problem in a computationally efficient way by using a PROM. The pipeline results in an optimal shape that outperforms the nominal setting.}
    \label{fig:visual_abstract}
\end{figure}

A typical way to simulate mechanical structures is to use the finite element method (FEM), which is based on the discretization of the original structure into finite elements (FEs). This approach, while expressive and accurate, often leads to high computational costs due to the large number of degrees of freedom (DoFs) of the resulting system. Moreover, these computational costs are notably increased when the system at hand is subject to nonlinear forces. In particular, large deformations of the structure lead to internal nonlinear forces, while contact with an external medium such as water or air gives rise to nonlinear drag forces. As a consequence, these large computational efforts might hinder the rapid prototyping and designing of new mechanical systems. 

To reduce the computational costs of FEM simulations, model order reduction techniques are widely used. In such cases, the use of a reduced order model (ROM) allows for faster simulations as it uses significantly less DoFs compared to the initial full order model (FOM). If properly chosen, the selected DoFs used in the ROM are the most significant for the modeling of the studied structure, and, therefore, the simulations remain accurate while being faster. In this context, different approaches can be followed to select a reduced order basis (ROB) that can correctly capture the deformation of the original nonlinear system. In addition, it is even possible to include parameters describing important characteristics of the considered mechanical structure in the ROM, resulting in a parametric reduced order model (PROM). The key advantage of a PROM is that it only needs to be constructed once and remains valid for multiple, small variations in the system's parameters.

\MD{Building on this particularity, which enables gradient computation, we propose a methodology for embedding nonlinear internal and external forces into a PROM and to exploit its analytical gradients to perform design optimization.} Building on our recent work using a tensorial approach to express internal nonlinear forces in PROMs~\citep{Marconi2021AExpansion,Marconi2020,Saccani2022}, we expand this method to reduce other (external) nonlinear forces that are relevant for the systems we study in this paper. \MD{We} leverage this approach to obtain well-structured analytical expressions of the various nonlinear forces acting on our system and embed their analytical gradients in a novel optimization pipeline. Because the nonlinear forces we consider show considerable diversity in their mathematical structure, we aim at providing a \MD{general} pipeline that can easily be adapted to other case-specific nonlinear forces.

To showcase our method, we consider the task of finding optimal shapes of soft robotic fishes (\Cref{fig:visual_abstract}). This choice is motivated by three important characteristics of soft robots that make them suitable benchmarks for our modeling and optimization approaches: (1) soft robots are highly deformable, resulting in nonlinear internal elastic forces that can be accurately modeled using FEM-based approaches, (2) soft swimmers are subject to diverse nonlinear external forces which slow down simulations and create a need for ROMs to capture the dynamics more efficiently, and (3) finding optimal shapes is of practical relevance for roboticists and is known as a difficult design task. Our contribution can therefore be summarized as follows: 
\MD{
\begin{enumerate}
    \item \textit{Development and testing of a fast optimization pipeline}: Our optimization pipeline efficiently leverages the gradients of our PROM to find optimal design shapes. By showcasing, comparing, and discussing several numerical case studies, we show the benefits and limitations of our algorithm.
    \item \textit{Creation of a specific PROM for nonlinear hydrodynamic forces}: Our PROM contributes to related work by extending existing reduction methods to new external forces. In particular, the hydrodynamic forces expressed by Lighthill's elongated-body theory are considered~\citep{lighthill1971large}. Our work provides examples for the reduction of different nonlinear forces that can serve as a blueprint for other studies.
\end{enumerate}}

The remainder of this paper is structured as follows. After highlighting relevant related work in~\Cref{sec:02_related_work}, we introduce the different nonlinear forces used in our framework in~\Cref{sec:03_force_modeling}. These forces are first expressed at the finite element level and then reduced in~\Cref{sec:04_PROM_section} to obtain our PROM formulation. We then present our optimization pipeline in~\Cref{sec:05_optimization_pipeline}, and show our numerical results in~\Cref{sec:06_results}. We conclude this work by discussing the relevance of our numerical approach, its limitations, and important directions for future work in~\Cref{sec:07_conclusion_future_work}.
\section{Related work}\label{sec:02_related_work}

This section reviews the most relevant literature related to the methods and results presented in this paper. It is divided into three parts. First, we discuss previous work important for the simulation of soft robots, and in particular for the simulation of soft swimmers. Second, we present literature on FEM-based ROMs, which can be leveraged to reduce the computational burden associated with FEM simulations. As a third and final part, we discuss the task of optimizing the shape of soft swimmers, and highlight how this task relates to both FEM and ROM-based simulations.

\subsection{Simulation of soft robotic fishes}

In comparison to rigid robots, soft robots are difficult to model and simulate due to the infinite number of degrees of freedom of their state space and the large nonlinear deformations they undergo~\citep{chen2023morphological, laschi2012soft, trivedi2008geometrically}. Methods to simulate soft robots can be divided into three main categories based on the type of models used to describe the robot dynamics. The first type of models consists of simplified models, such as the augmented rigid body model~\citep{della2020improved,della_santina_model-based_2020}, which is based on the piecewise constant curvature model~\citep{webster2010design}. A second type of models relies on data collected on the robot itself~\citep{huang2024high,bruder2025koopman, bruder2019modeling,holsten2019data, reinhart2016hybrid}. Finally, a third type of models considers FEM as a way to simulate soft robots~\citep{wang2024fin, qin2024modeling,Du2021,sofa,duriez2013control}. This last type of models is considered in this work.

At their core, FEM approaches follow a few key steps: they discretize the structure at hand in small, finite elements (FEs), formulate equations for each FE considering the different forces applied to the structure, assemble the FE into a global system, and solve the resulting set of equations to describe quantities such as displacement over time. Following these steps, FEM approaches can effectively account for nonlinearities that arise from the large deformations of soft robots and the nonlinear behavior of the materials used to manufacture these robots~\citep{tawk2020finite, pinskier2024diversity}. In practice, however, achieving precise FEM-based simulations requires careful meshing and precise tuning of material parameters~\citep{dubied2022sim}. Even then, high-fidelity nonlinear FEM simulations are computationally expensive without model order reduction techniques~\citep{navez2025modeling,goury2021real}. Therefore, simulations using FOMs are not directly suitable for control and fast prototyping tasks. 

Before discussing model order reduction strategies in \Cref{subsec:related_work_ROM_FEM}, we first focus on the specific modeling of soft robotic fishes,  which are the mechanical structures that we study in this work. Due to their agile and efficient underwater movements, real fishes have inspired roboticists aiming to replicate these desirable characteristics in soft robotic designs~\citep{katzschmann_hydraulic_2016, zhu_tuna_2019, van_den_berg_biomimetic_2020, wang2024fast, liu2025design}. Designing such robotic fishes is time-consuming and requires iterative fabrication and testing procedures. In this context, numerical simulations can fasten the process by enabling virtual design iterations and testing~\citep{matthews2023efficient,spielberg2017functional}. Nevertheless, using FEM-based simulations for soft robotic fishes presents significant challenges, particularly due to the complexities of fluid-structure interaction (FSI) in underwater environments~\citep{nava_fast_2022}. Due to this complexity, previous works modeling FSI using FEM typically restrict their attention to planar formulations~\citep{curatolo2015virtual,gravert2022planar,wang2024novel} or use parametrized added-mass and drag forces~\citep{liu2025design}.

When moving from 2-dimensional to 3-dimensional fish FEM simulations, the effects of hydrodynamics on the solid structure are often modeled using surrogate models~\citep{SoftCon,diffaqua}. Practically, these models aim to approximate the thrust and drag caused by the fluid on the robot using heuristic rules. Alternatively, neural networks~\citep{wandel2020learning,zhang2022sim2real} and GPU-optimized solvers~\citep{liu2022fishgym} have been used to simulate fluid-solid interactions. In this work, we consider Lighthill's large-amplitude elongated-body theory (LAEBT) to model the interaction between the fluid and the fish movements~\citep{lighthill1971large}. The LAEBT models the thrust (or reactive) force produced by the locomotion of carangiform fish, considering the rate of change in momentum in the fluid domain surrounding the fish. The LAEBT, as well as its extensions, have been compared to computational fluid dynamics (CFD) simulations~\citep{candelier2011three,candelier2013note} and real-world experiments~\citep{wang2011dynamic,li2014modeling, chen2023experimental}. As a key contribution, we show in this work how to reduce the hydrodynamic forces stemming from the LAEBT into a PROM, enabling the faster solving of equations of motion.

\subsection{ROMs for FEM-based simulations}\label{subsec:related_work_ROM_FEM}

Model order reduction (MOR) techniques have been developed to mitigate the computational costs associated with FEM simulations, especially when dealing with nonlinear and high-dimensional systems. In this context, reduced-order models (ROMs) provide an efficient way to approximate the behavior of complex mechanical systems while retaining sufficient accuracy for practical applications.

For systems with geometric and material nonlinearities, tensor-based formulations have recently emerged as powerful tools to represent internal forces in a polynomial and structured form. In particular, \cite{Marconi2020,Marconi2021AExpansion} introduced a nonlinear ROM framework based on a Neumann expansion of the deformation gradient, allowing internal elastic forces to be expressed as high-order tensors. This structure enables efficient evaluation and differentiation, which is essential for gradient-based optimization. The method has been further extended to include shape defects and parametric variations, thus broadening its applicability. Along with allowing analytical parametrization with respect to shape variations, the tensorial approach also avoids the bottleneck of the computation of the reduced nonlinear terms and therefore does not require hyper-reduction, as done for instance in \cite{Tiso2013}. 

\cite{Saccani2022} built upon this approach to perform sensitivity analysis of nonlinear frequency responses in structures with geometric imperfections. Their work demonstrated how the parametric ROM formulation can be used for fast and accurate computation of nonlinear dynamic responses under varying conditions. For multibody flexible systems, the mentioned approaches could embedded into component mode synthesis methods \citep{KARAMOOZMAHDIABADI2019114915,Wu2019,WuTiso2016}.

ROMs have also proven valuable for soft robotics, where nonlinearities arise from both the material behavior and interactions with the environment. For example, \cite{Goury2018} proposed a data-driven MOR approach tailored to soft robots, enabling real-time control and simulation. \cite{Katzschmann2019} applied ROMs in the context of soft robotic arms, using a reduced FEM model with a state observer for closed-loop control. More recently, \cite{navez2025modeling} proposed a MOR method for soft robots with contact-rich dynamics and a variety of actuators, further confirming the effectiveness of ROM-based approaches for simulation and control in such settings.


\subsection{Shape optimization of soft swimmers}

Equipped with reliable simulation frameworks, roboticists can design soft robots efficiently by first analyzing virtual designs, and in a second step proceed to the time-consuming manufacturing of these designs. Going one step further, differentiable simulators provide their users with gradients, allowing therefore the use of automated gradient-based optimization~\citep{Bacher2021DesignSimulation}.

Various differentiable frameworks suitable for the simulation of soft robots have recently emerged~\citep{diffpd,geilinger2020,chainqueen}. Previous works allow for the co-design of structural and control parameters for situations with contact forces, but without using any ROM formulation~\citep{diffpd,geilinger2020,chainqueen,hahn_real2sim_2019,hu_difftaichi_2020, huang_plasticinelab_2021, bern2019trajectory, wang2024fin}. As an example, the DiffPD differentiable simulator~\citep{diffpd} allows the automatic calibration of material parameters and control signals applied to soft structures modeled with FEM. The recent work by \cite{navez2025modeling} goes one step further by presenting a data-driven MOR method that can be used to optimize control and design parameters. 
The proposed method uses a neural network to learn a representation of soft robots based on a large dataset of FOM FEM simulations. In contrast, our pipeline does not require any prior FEM simulation or offline learning to construct the PROM, and we focus on a larger set of design parameters.


The work most similar to ours in terms of application is probably the DiffAqua pipeline developed by~\cite{diffaqua}, which employs a differentiable framework to optimize the design of soft swimmers. It uses the Wasserstein barycenters to interpolate between different base shapes and find an optimal shape combination, using a FOM. On a methodological level, our optimization method is similar to the one proposed by \cite{frohlich2019geometric}, where a PROM is used to optimize shape parameters for linear systems. In comparison, we present a method that allows to include nonlinear forces in the PROM and the optimization pipeline. \MD{Importantly, our starting point is to express internal and external forces using a full FE mesh and only then project them onto a reduced model for optimization, in contrast to approaches that formulate the optimization directly on low-order models such as Cosserat rod formulations \citep{trivedi2008optimal, armanini2022model}.}

Our work combines the different key elements mentioned above: shape optimization, data-free PROM, differentiable framework, and hydrodynamic forces acting as nonlinear forces. By combining these elements, we develop in this paper a fast and accurate optimization algorithm to find optimal soft swimmer shapes.

\section{Polynomial forces at the finite element-level}\label{sec:03_force_modeling}

In this section, we describe the forces we include in our FE model. After presenting the equations of motion (EoMs) of the FOM in~\Cref{sec:nonlinear_FEM}, we derive polynomial expressions for the forces acting on the soft swimmers: the nonlinear internal forces (\Cref{sec:internal_forces}), the thrust force (\Cref{sec:thrust_force}), the drag force (\Cref{sec:thrust_force}) and the actuation force (\Cref{sec:actuation_force}). These forces, expressed at the element-level, are then reduced and included in the PROM in~\Cref{sec:04_PROM_section}.

\subsection{Nonlinear full order model}\label{sec:nonlinear_FEM}
The nonlinear FE system we consider in this work is
\begin{align}\label{eq:EoMs_nonlinear}
    \M \udd +\C \udot + \fint(\disp) 
    &= 
    \ft(\disp,\udot,\udd) 
    \nonumber
    \\
    &
    + \fd(\udot) + \fact(\disp,t),
\end{align}
where~$\disp\in\mathbb{R}^{n}$ is the nodal displacement vector,~$\M$ the mass matrix,~$\C$ the damping matrix, and~$\fint$ the internal nonlinear forces. The external forces acting on the fish are separated into the thrust force~$\ft$, the drag force~$\fd$ and the actuation force~$\fact$. In the following, we denote vectors by bold lower case letters and matrices as well as tensors by bold upper case letters.

The FEM formulation used in this work includes a set of user-defined parameters that allows to describe shape variations of the original, \textit{nominal} structure. These shape variations (see, e.g.,~\Cref{fig:visual_abstract}, and~\Cref{fig:all_shape_variations}) describe alternative FE meshes which we call \textit{shape-varied} configurations. In our optimization framework (\Cref{sec:05_optimization_pipeline}), the linear combinations of these shape-varied configurations defines our search space. 

\begin{figure}[tb]
    \centering
    \includegraphics[width=\linewidth]{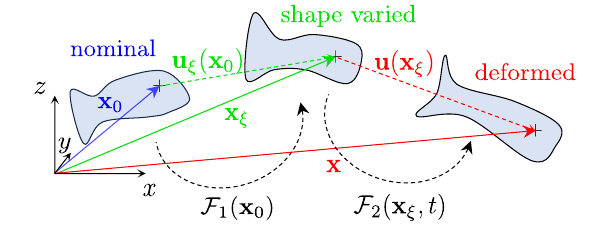}
    \caption{Nominal, shape-varied, and deformed configurations. The final deformed configuration can be expressed through two successive mappings:~$\mathcal{F}_1(\xo)$ and~$\mathcal{F}_2(\x_\xi,t)$.}
    \label{fig:deformation_setting}
\end{figure}

Following the approach presented by~\cite{Marconi2021AExpansion}, the deformations of shape-varied meshes are described by two successive mappings, as shown in~\Cref{fig:deformation_setting}. The first mapping,~$\mathcal{F}_1(\x_0)$, describes the mesh deformation from the nominal configuration~$\x_0$ to the shape-varied configuration~$\uxi$, and the second mapping,~$\mathcal{F}_2(\x_\xi,t)$, the deformation from the shape-varied configuration to the deformed configuration~$\x$. This 2-steps approach is used to describe the different forces acting on the FE mesh, which we derive next.

\subsection{Nonlinear internal forces}\label{sec:internal_forces}

The nonlinear internal forces are formulated according to the approach presented by \cite{Marconi2021AExpansion}. Using the notation~$\D = \frac{\partial\bm{u}}{\partial\xo}$ and~$\D_\xi = \frac{\partial\bm{u}_\xi}{\partial\xo}$, together with the deformation gradients~$\F_1$ and~$\F_2$ corresponding to the mappings~$\mathcal{F}_1$ and~$\mathcal{F}_2$ respectively, the Green-Lagrange strain can be expressed as
\begin{align*}
    \E
    &= \frac{1}{2}(\F_2^\top\F_2- \I)
    \\
    &= \frac{1}{2}\F_1^{-\top}(\D+\Dt+\Dt\D+ \Dxit\D+\Dt\Dxi)\F_1^{-1}.
    \nonumber
\end{align*}

The Green-Lagrange strain can be expressed as a polynomial by approximating the inverse deformation gradient~$\F_1^{-1}$ using convergent Neumann series~\citep{Marconi2020,Marconi2021AExpansion}. This polynomial form of the Green-Lagrange strain implies that the internal force~$\fint$ is also polynomial in~$\disp$ and~$\ud$. At the element-level of the FE assembly, the internal force is composed of three separate contributions:
\begin{equation*}
    \fint^e = \f_1^e + \f_2^e + \f_3^e \in\mathbb{R}^{n_e},
\end{equation*}
where the superscript~$\star^e$ denotes element-level quantities and~$n_e$ is the number of DoFs in a single FE. These three contributions are given by
\begin{align*}
    \f_1^e &= \ps{}{2}\K (\uxi^e)\cdot\disp^e,\\
    \f_2^e &= \ps{}{3}\K (\uxi^e):(\disp^e\otimes\disp^e),\\
    \f_3^e &= \ps{}{4}\K (\uxi^e)	\:\vdots\: (\disp^e\otimes\disp^e\otimes\disp^e),
\end{align*}
where 
\begin{subequations}\label{eq:tensors_internal_force}
    \begin{align}
        \ps{}{2}\K(\disp_d^e)   
        &
        = \ps{}{2n}\K + \ps{}{3\xi}\K\cdot\uxi^e + \ps{}{4\xi\xi}\K:(\uxi^e\otimes\uxi^e),
        \\
        \ps{}{3}\K(\disp_d^e)   
        &
        = \ps{}{3n}\K +\ps{}{4\xi}\K\cdot\uxi^e + \ps{}{5\xi\xi}\K:(\uxi^e\otimes\uxi^e),
        \\
        \ps{}{4}\K(\uxi^e)   
        &
        = \ps{}{4n}\K +\ps{}{5\xi}\K\cdot\uxi^e + \ps{}{6\xi\xi}\K:(\uxi^e\otimes\uxi^e).
    \end{align} 
\end{subequations}
The subscripts of the tensors~$\K$ should be understood as follows:
\begin{itemize}
    \item The number denotes the order of the tensor.
    \item The letter~$n$ denotes the fact that the tensor is computed for the nominal mesh.
    \item The letter~$\xi$ denotes the fact that the tensor multiplies the shape-varied vector~$\uxi$, either once ($\xi$) or twice ($\xi\xi$).
\end{itemize}
The exact formulation of these tensors can be found in \cite{Marconi2021AExpansion}.


\subsection{Thrust force}\label{sec:thrust_force}

The interaction between the fluid and the fish is modeled using the Lighthill's elongated-body theory of fish locomotion \citep{lighthill1971large}. While being a simplification of the complex solid-fluid interactions that can happen in reality, this model is derived using physics' first principles. In addition, this theory has been used to analyze the locomotion of fishes and underwater robots in practice~\citep{wang2011dynamic, li2014modeling, diffaqua, eloy2024flow}. 

\begin{figure}[tb]
    \centering
    \includegraphics[width=\linewidth]{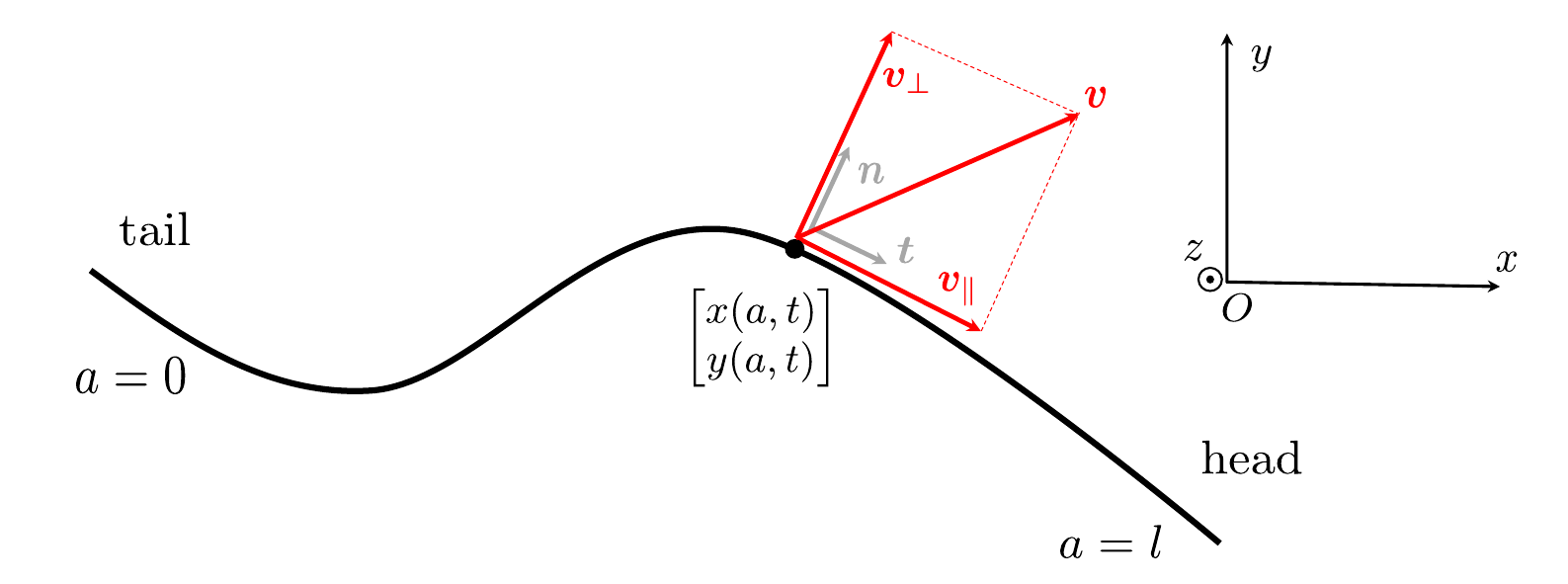}
    \caption{Reference frame used to derive the reactive force. The spine of the fish is observed from the top in this diagram.}
    \label{fig:lighthill_elongated_body}
\end{figure}

\Cref{fig:lighthill_elongated_body} introduces the reference frame used to describe the hydrodynamic forces acting on the fish. Particularly important is the use of the Lagrangian coordinate~$a$ which has a value of~$0$ at the tail of the fish, and~$l$ at its head.

The Lighthill's elongated-body theory of fish locomotion derives a force called the \textit{reactive force}, which captures the effects of inertial forces on the fish locomotion in inviscid flows. This reactive force, responsible for the thrust, is given by
\begin{align}\label{eq:reactive_force}
    \begin{bmatrix}f_x\\f_y\end{bmatrix}
    =
    &
    \begin{bmatrix}-\mt v_{\parallel}v_{\perp}\bm{n}+\frac{1}{2}\mt v_{\perp}^2\bm{t}\end{bmatrix}_{a=0}
    \nonumber
    \\
    & 
    -\frac{\text{d}}{\text{d}t}\int_{0}^{l}\mt v_{\perp}\bm{n}\;\text{d}a,
    \nonumber
    \\
    =
    & \;\ftail + \fspine
\end{align}
where~$\bm{t}$ is the unit vector tangential to the spine pointing in the direction of the head, and~$\bm{n}$ is the unit vector normal to the spine (see \Cref{fig:lighthill_elongated_body} for the other symbols). In addition,~$\mt$ is the virtual mass of the considered cross-section of the fish. According to  \cite{lighthill1971large}, this virtual mass is captured by the expression
\begin{equation}\label{eq:virtual_mass}
    \mt = \frac{\pi}{4}\rho d^2,
\end{equation}
where~$\rho$ is the density of the fluid and~$d$ the cross-section's depth in the~$z$-direction. 

In the following, we denote the first part of the reactive force~\eqref{eq:reactive_force} as the \textit{tail force} (at~$a=0$) and the second part as the \textit{spine force}, which we combine to obtain the \textit{thrust force}~$\ft$ in the EoMs~\eqref{eq:EoMs_nonlinear}. Moreover, we assume that at the tail ($a=0$), the perpendicular velocity~$v_{\perp}$ is much larger than the tangential velocity~$v_{\parallel}$~\citep{wang2011dynamic}. Next, we formulate the tail and spine force at the element-level of the FE model, after defining which nodes of the FE mesh are subject to these forces. When not necessary, we omit the superscript~$\star^e$ to present more readable expressions.

\subsubsection{Selection of nodes} 

As~\cite{lighthill1971large} derives the reactive force by focusing on the spine of the fish, we select \textit{spine nodes} in the FE mesh, to which the reactive force~\eqref{eq:reactive_force} is applied. We rely on a symmetric meshing of the nominal structure, which therefore contains clearly defined spine nodes located on the symmetry axis. For each successive pair of spine nodes, we attribute a single FE which we include in a set of \textit{spine elements}. These spine elements are then used to apply the tail and spine forces, as described below. 

\subsubsection{Tail force} The tail force is applied at the tail node, which is part of the spine element located at the extremity of the fish ($a=0$). At this element, and with the assumption that~$v_{\perp}\gg v_{\parallel}$~\citep{wang2011dynamic}, the tail force is
\begin{align}\label{eq:tail_force}
    \ftail^e 
    &=\frac{1}{2}\mt v_{\perp}^2\bm{t}^e
    = \frac{1}{2}\mt[(\bm{v}\cdot\bm{n}^e)^2\bm{t}^e]
    \nonumber
    \\
    &= \frac{1}{2}\mt w^3 [(\A\udot)^\top \R \B\x]^2\B \x,
\end{align}
with $\x = \xo+\uxi+\disp$ (see \Cref{fig:deformation_setting}) and where~$\bm{t}^e,\bm{n}^e\in\mathbb{R}^{n_e}$ are the element versions of the 3D tangential vectors~$\bm{t}$ and~$\bm{n}$. In addition, the matrix~$\A$ selects the component of~$\udot$ corresponding to the tail DoFs,~$\B$ is used to create the tangential vector~$\bm{t}^e$, and~$\R$ is a rotation matrix. The factor~$w$ stems from the normalization of the vectors~$\bm{t}^e$ and~$\bm{n}^e$, assuming that the length of the spine is constant. 

\subsubsection{Spine force} The spine force is applied to each spine element. For a given spine FE, the force at the element-level is described as
\begin{alignat}{2}\label{eq:spine_force}
    \fspine^e 
    &
    = -\der{}{t}\int_0^l &&\mt (\A\udot)^\top \R\B\x w \R\B \x w\text{d}a
    \nonumber
    \\
    &
    = -\mt w \der{}{t} &&\left[(\A\udot)^\top \R\B\x \R\B \x\right]
    \nonumber
    \\
    &
    = -\mt w &&\left[(\A\udd)^\top \R\B\x \R\B \x\right.
    \nonumber
    \\
    & &&+(\A\udot)^\top \R\B\udot \R\B\x
    \nonumber
    \\
    & && \left.+(\A\udot)^\top \R\B\x \R\B \udot\right].
\end{alignat}


\subsubsection{Virtual mass} The virtual mass~$\mt$~\eqref{eq:virtual_mass} of a given cross-section of the fish depends on the considered cross-section depth~$d$. For each spine element, we match a single dorsal node with~$z$-coordinate~$z_0^{\mathrm{max}} + z_\xi^{\mathrm{max}}$. This allows us to compute the virtual mass attributed to a given spine element as
\begin{equation*}
    \mt = \frac{\pi}{4}\rho d^2 =\frac{\pi}{4}\rho (2z_0^{\mathrm{max}}+ 2z_\xi^{\mathrm{max}})^2.
\end{equation*}

\subsection{Drag force}\label{sec:drag_force}

While the reactive force~\eqref{eq:reactive_force} derived by \cite{lighthill1971large} accounts for the inertial interaction between the fluid and the fish, it considers inviscid conditions where no drag occurs. We therefore enhance our model with a form drag based on the shape of the fish. Specifically, we apply a force at each element located on the skin of the fish similarly to the approach of~\cite{diffaqua} and \cite{SoftCon}. The drag force depends on the orientation of the surface elements and the overall fish velocity in the forward swimming direction ($x$-direction):
\begin{equation*}
    \fd = -\frac{1}{2}\rho A C_d(\alpha) ||\bm{v}_x||^2\dvec_{\mathrm{swim}},
\end{equation*}
where~$A$ is the area of the skin surface,~$C_d$ is a drag coefficient depending on the angle of attack~$\alpha$ of the considered surface, and~$\dvec_{\mathrm{swim}}$ is a unit vector pointing in the forward swimming direction. The velocity~$\bm{v}_x$ is measured at the fish head. The drag coefficient is symmetric with respect to~$\alpha$ and expressed by analyzing the shape-varied configuration at rest as 
\begin{align*}
    C_d(\alpha) 
    & = \cos(2\alpha -\pi)+1
    \nonumber
    \\
    &=  \cos(2\arccos(\nv\cdot \dvec_{\mathrm{swim}})-\pi+\pi)+1
    \nonumber
    \\
    &=  2(\nv\cdot\dvec_{\mathrm{swim}})^2.
\end{align*}

In contrast to the tail force~\eqref{eq:tail_force} and the spine force~\eqref{eq:spine_force}, the drag force is not a polynomial in~$\uxi$. As a polynomial force is required to build our PROM (see~\Cref{sec:04_PROM_section}), we approximate the drag force using a Taylor approximation up to the third order around the nominal structure~$\uxi=\bm{0}$. The Taylor series coefficients are in this case tensors~$\T$ whose analytical form can be found using a computer algebra system (CAS) offline (Mathematical in our case). This results in the following drag force at the element-level:
\begin{equation}\label{eq:drag_force_taylor}
        \fd^e = 
         \big(\prescript{}{3}\T + \prescript{}{4}\T \cdot \uxi 
        + \prescript{}{5}\T : (\uxi\otimes\uxi) \big):(\udot \otimes \udot).
\end{equation}


\subsection{Actuation force}\label{sec:actuation_force}

The actuation force~$\fact$ is inspired by the muscle model used by \cite{diffpd} and \cite{SoftCon}. This approach allocates an additional spring-like energy to specific elements of the mesh, called \textit{muscle elements}, which together form muscles that can contract or extend. We allocate the following energy to the muscle elements:
\begin{equation}\label{eq:muscle_energy}
    E = \frac{k}{2}aV^e||\F \cdot\bm{m}||^2,
\end{equation}
where~$k$ can be understood as a spring stiffness,~$a$ is an actuation signal, $V^e$ is the volume of the element, ~$\F$ is the deformation gradient, and~$\bm{m}$ is the direction of actuation defined for the undeformed mesh element. Extension occurs for~$a<0$, and contraction occurs for~$a>0$. For simplicity, we consider $V^e$ as a constant for each element and compute it based on the nominal configuration of the fish. Limitations of this approach are further discussed in \Cref{sec:multi_obj_co_opti}.

Before deriving an expression for the actuation forces~$\fact$ from the energy \eqref{eq:muscle_energy}, we express this energy in a convenient tensorial form. Using a first order Neumann expansion as in \cite{Marconi2021AExpansion}, we can approximate the deformation gradient as 
\begin{equation*}
    \F  = \F_2\F_1\approx \I + \Dxi + \D - \D\Dxi^2.
\end{equation*}
Neglecting the~$\mathcal{O}(\D_d^2)$ terms, we get
\begin{equation}\label{deformation_gradient_actuation}
    \F \approx \I + \Dxi + \D.
\end{equation}
The muscle energy~\eqref{eq:muscle_energy} can be written using the obtained deformation gradient~\eqref{deformation_gradient_actuation}:
\begin{align}\label{eq:muscle_energy_extended}
    E   &=  \frac{k}{2}a V^e \m^\top \F^\top \F \m                            
        \nonumber
        \\
        &=  \frac{k}{2}a V^e \m^\top (\I+\D+\D^\top + \D^\top\D \nonumber
        \nonumber
        \\
        &+ \Dxi+ \Dxi^\top + \Dxi^\top \Dxi + \Dxi^\top \D + \D^\top\Dxi)\m .
\end{align}
We note that the matrix described by the expression in the parentheses is symmetric. At the element-level, we can resort to the shape function derivatives matrix~$\G \in \mathbb{R}^{9\times n_e}$ to express the terms in \eqref{deformation_gradient_actuation}. This can be done using the Voigt notation, which we denote with a subscript~$\star_V$.  We first express the matrices of \eqref{eq:muscle_energy_extended} using in the Voigt notation:
\begin{align*}
     \I     &\longleftrightarrow     \I_V\\
     \D+\D^\top + \D^\top\D &\longleftrightarrow (2\Hm+ \A_1(\thetav))\thetav \\
     \Dxi+\Dxi^\top + \Dxi^\top\Dxi &\longleftrightarrow (2\Hm+ \A_1(\thetav_\xi))\thetav_\xi \\
     \Dxi^\top\D+\D^\top\Dxi &\longleftrightarrow 2\A_1(\thetav_\xi)\thetav,
\end{align*}
where~$\bm{\theta}$ and~$\bm{\theta}_d$ are the vectorized forms of~$\D$ and~$\D_d$, e.g.,~$\bm{\theta}= [u_{,x}\:u_{,y}\:u_{,z}\:v_{,x}\:v_{,y}\:v_{,z}\:w_{,x}\:w_{,y}\:w_{,z}\:]^\top$. The matrices~$\I_V,\Hm$,$\A_1(\thetav)$,~$\A_1(\thetav_\xi)$ are given in \Cref{A_actuation_force}.

The Voigt notation allows us to express the muscle energy as
\begin{equation}\label{eq:muscle_energy_voigt}
\begin{aligned}
E_V &= \frac{k}{2}a V^e \m_V^\top \left[\I_V + (2\Hm+ \A_1(\thetav))\thetav \right.
\\
&\left. +(2\Hm+ \A_1(\thetav_\xi))\thetav_\xi + 2\A_1(\thetav_\xi)\thetav \right],
\end{aligned}
\end{equation}
with~$\m_V = [m_x^2, m_y^2, m_z^2, m_xm_y, m_xm_z, m_ym_z]^\top$.
We can use the shape function derivatives~$\G$ to express \eqref{eq:muscle_energy_voigt} as an explicit function of the displacement~$\disp$ at the element-level:
\begin{align*}
     E_V 
     &= \frac{k}{2}a V^e\m_V^\top \left[\I_V + (2\Hm+ \Lm_1\cdot\thetav)\thetav\right.\nonumber \\
     &+\left.(2\Hm+ \Lm_1\cdot\thetav_\xi)\thetav_\xi + 2\Lm_1\cdot\thetav_\xi\thetav \right]
     \\
     &= \frac{k}{2}a \m_V^\top \left[\I_V + (2\Hm+ \Lm_1\cdot\G\disp)\G\disp\right. \nonumber\\
     &+\left. (2\Hm+ \Lm_1\cdot\G\uxi)\G\uxi + 2\Lm_1\cdot\G\uxi\G\disp \right],
\end{align*}
where~$\Lm_1$ is a sparse third order tensor. Its exact composition is given in \Cref{A_actuation_force}. The operation between the tensor~$\Lm_1$ and the vectors~$\G\disp$ and~$\G\uxi$ is a contraction, which is best expressed using Einstein's notation:
\begin{align*}
    E_V &= \frac{k}{2}a V^e m_{V,i} \left[I_{V,i} + (2H_{ij}+ L_{1,ijk}G_{kl} u_l)G_{jm} u_m\right. \\
     &+\left. (2H_{ij}+ L_{1,ijk}G_{kl} u_{\xi,l})G_{jm} u_{\xi,m} \right.
     \\
     &\left. + 2L_{1,ijk}G_{kl}u_{\xi,l}G_{jm}u_m \right].
\end{align*}

Finally, we can take the derivative of the muscle energy to get the corresponding actuation forces at the element-level, as follows:
\begin{equation}\label{eq:actuation_force}
    \begin{aligned}
        f^e_{\text{act},L}
        &=\der{E_V}{u_L} 
        \\
        &=\frac{k}{2}a V^e m_{V,i} \bigg[(2H_{ij}+ L_{1,ijk}G_{kl} u_l)G_{jL}
        \\
        &+ L_{1,ijk}G_{kL}G_{jm}u_m+  2L_{1,ijk}G_{kl}u_{d,l}G_{jL} \bigg],
    \end{aligned}
\end{equation}
where~$f^e_{\text{act},L}$ is the component~$L$ of the actuation force~$\fact$ at the element-level.

\section{Reduction of nonlinear forces in PROM}\label{sec:04_PROM_section}

In this section, we present our PROM, which allows to obtain a reduced order version of the EoMs~\eqref{eq:EoMs_nonlinear}, i.e., 
\begin{align}\label{eq:red_EoMs}
    \M^r \etadd +\C^r \etad + \fint^r(\etav) 
    &= 
    \ft^r(\etav,\etad,\etadd) 
    \nonumber
    \\
    &
    + \fd^r(\etad) + \fact^r(\etav,t),
\end{align}
where~$\etav\in \mathbb{R}^m$ entails the DoFs of the reduced system with dimension~$m\ll n$, and the superscript~$r$ denotes reduced quantities. To obtain these reduced quantities, the contribution of each element-level force is projected onto a ROB through a Galerkin projection. After presenting how we build our ROB (\Cref{sec:PROM_ROB}), we derive the reduced version of the forces introduced in \Cref{sec:03_force_modeling} at the element-level (\Cref{sec:reduced_internal_force},~\ref{sec:reduced_thrust_force},~\ref{sec:reduced_drag_force}, and~\ref{eq:reduced_actuation_force}). The reduced element-level objects (e.g., tensors) can then simply be summed to obtain the EoMs~\eqref{eq:red_EoMs} associated with the PROM.

\subsection{Reduced order basis}\label{sec:PROM_ROB}

\begin{figure}[tb]
    \centering
    \includegraphics[width=\linewidth]{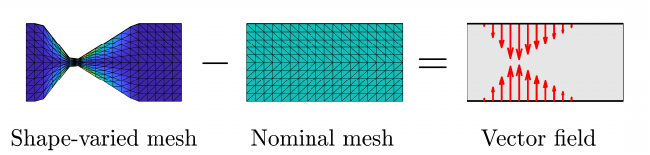}
    \caption{Procedure for populating the matrix~$\U$ used to describe the shape-varied configuration~$\uxi = \U\xiv$. After defining a shape-varied mesh, we subtract the nominal node positions to obtain a vector field. This vector field is stored as a column of~$\U$. By defining multiple vector field in the different columns of~$\U$, we can scale and combine them by multiplying~$\U$ with the parameter vector~$\xiv$ to obtain the different shape-varied configurations~$\uxi$.}
    \label{fig:shape_variations_procedure}
\end{figure}

The~$m$ basis vectors of the reduced order basis (ROB), which we describe in this section, are grouped in the matrix~$\V\in\mathbb{R}^{n\times m}$. The ROB described by~$\V$ can be used to approximate the FOM displacement~$\disp$ using the reduced order displacement~$\etav$ as~$\disp \approx \V\etav$. Similarly, a basis can be created for the user-defined~$m_\xi$ shape variations as~$\uxi = \U\xiv$ with~$\U\in\mathbb{R}^{n\times m_\xi}$. We refer to the vector~$\xiv$ as \textit{parameter vector}. \Cref{fig:shape_variations_procedure} shows how the matrix~$\U$ is constructed by concatenating multiple basis vectors describing different shape variations. These shape variations are linearly combined to obtain the shape-varied configuration, through the operation~$\uxi=\U\xiv$.

Following the work of~\cite{Marconi2020, Marconi2021AExpansion,Saccani2022}, we use a combination of vibration modes (VMs), modal derivatives (MDs), as well as shape variation parameter sensitivities (PSs) as basis vectors to construct our ROB. The VMs, MDs and PSs are stored in the matrices~$\bm{\Phi}$,~$\bm{\Theta}$ and~$\bm{\Xi}$, respectively, and are assembled in~$\V = [\bm{\Phi},\bm{\Theta}, \bm{\Xi}]$ to obtain the ROB. To improve the numerical properties of the ROB, we orthonormalize~$\V$.

The VMs are obtained by considering the following linearized undamped system:
\begin{equation}\label{eq:linear_undamped_system}
    \M \udd +\K\disp = \ze,
\end{equation}
\MD{where $\K$ is the tangent stiffness matrix.} The solutions to the differential equation~\eqref{eq:linear_undamped_system} can be found by solving the related eigenvalue problem
\begin{equation}\label{eq:eigenvalue_problem_VM}
    (\K-\omega_i^2\M)\phiv_i = \ze.
\end{equation}
where~$\omega_i$ is the~$i$-th eigenvalue, and~$\phiv_i$ is the~$i$-th eigenvector, which corresponds to the~$i$-th column of~$\bm{\Phi}$.

The MDs incorporated in the ROB are computed from the derivative of~\eqref{eq:eigenvalue_problem_VM} with respect to the displacement~$\disp$:
\begin{equation*}
     (\K-\omega_i^2\M)\pop{\phiv_i}{u_j}+ \left(\pop{\K}{u_j}-\pop{\omega_i}{u_j}\right)\phiv_i = \ze.
\end{equation*}
The enhancement of the ROB with the MDs~$\theta_{ij}=\pop{\phiv_i}{u_j}$ allows to capture part of the effects of the nonlinear system. The quantities~$\theta_{ij}$ are the element of the matrix~$\bm{\Theta}$.

Finally, the PSs are obtained in a similar way as the MDs, but this time considering the differentiation of~\eqref{eq:eigenvalue_problem_VM} with respect to the shape variation parameter vector~$\xiv$:
\begin{equation*}
     (\K-\omega_i^2\M)\pop{\phiv_i}{\xi_j}+ \left(\pop{\K}{\xi_j}-\pop{\omega_i}{\xi_j}\right)\phiv_i = \ze.
\end{equation*}
Solving for~$\pop{\phiv_i}{\xi_j}$ allows to include the PSs in our ROB as elements of the matrix~$\bm{\Xi}$. \MD{Both the MDs and PSs are obtained in their static form computed on the nominal structure~\citep[see][Section 3.3 for details]{jain2017quadratic}.}

A notable advantage of this choice of ROB is the fact that its construction is data-free, meaning that we do not rely on any simulation of the FOM or observation of a real structure to obtain the PROM, which can be numerically and time-costly operations.

\subsection{Reduced internal forces}\label{sec:reduced_internal_force}

The reduced version of the internal force at the element-level consists of the projection of the tensors~$\K$~\eqref{eq:tensors_internal_force} onto the ROB. These reduced tensors are directly taken from previous work~\citep{Marconi2021AExpansion}.

\subsection{Reduced thrust force}\label{sec:reduced_thrust_force}

The reduced version of the thrust force, consisting of the tail force~\eqref{eq:tail_force} and the spine force~\eqref{eq:spine_force}, is obtained through a projection onto the ROB, as shown below.

\subsubsection{Tail force} The reduced version of the element-level tail force~\eqref{eq:tail_force} is
 \begin{align*}
      \ftail^r
    =
    &\frac{1}{2}\mt w^3 \left[(\A\V\etad)^\top \R \B(\xo+\U\xiv+\V\etav)\right]^2
    \\
    &\V^\top \B (\xo+\U\xiv+ \V\etav). 
 \end{align*}
The virtual mass is computed as 
\begin{equation*}
    \mt = \frac{\pi}{4}\rho d^2 = \frac{\pi}{4}\rho (2z_0^{\mathrm{max}}+2\U^{\mathrm{max}}\xiv)^2,
\end{equation*}
where~$\U^{\mathrm{max}}$ is the row of~$\U$ corresponding to the DOF of~$z_0^{\mathrm{max}}$.

\subsubsection{Spine force} The reduced version of the spine force~\eqref{eq:spine_force} is best described using Einstein's notation. The $I$-th component of the element-level force is given by
\begin{align}\label{eq:reduced_spine_force}
    f^r_I 
    = 
    - \frac{\pi}{4}\rho &
    (2z_0^{\mathrm{max}}+ 2U^{\mathrm{max}}_k\xi_k)^2 w 
    \\
    \cdot V_{iI} &
    \big\{  
    (A_{mn}V_{nj}\etaddnv_{j}) R_{mp}B_{pq} \x^r_{q}R_{is}B_{st}  \x^r_{t}
    \big. 
    \nonumber
    \\
    & 
    +(A_{mn}V_{nj}\etadnv_{j}) R_{mp}B_{pq}V_{qr}\etadnv_{r} R_{is}B_{st}  \x^r_{t}
    \big.
    \nonumber
    \\
    & 
    \Bigg.+(A_{mn}V_{nj}\etadnv_{j}) R_{mp}B_{pq} \x^r_{q} R_{is}B_{st} V_{tu}\etadnv_{u}\big\},
    \nonumber
\end{align}
where
\begin{equation*}
    \x^r_{q} = x_{0,q}+U_{qr}\xi_{r}+V_{qr}\eta_{r}.
\end{equation*}

To be able to sum the contribution of each spine element when building the PROM, it is necessary to express each coefficient of~\eqref{eq:reduced_spine_force} as separate tensors. These tensors are given in \Cref{A_reactive_force}.

\subsection{Reduced drag force}\label{sec:reduced_drag_force}

The reduced version of the drag force~\eqref{eq:drag_force_taylor} is given by
\begin{equation*}
        \fd^r = 
         \big(\prescript{}{3}\T^r + \prescript{}{4}\T^r \cdot \xiv 
        + \prescript{}{5}\T^r : (\xiv\otimes\xiv) \big) : (\etad \otimes \etad),
\end{equation*}
with, using Einstein notation,
\begin{align*}
    \ps{}{3}T^r_{IJK} &= V^e_{iI}\ps{}{3}T_{ijk}V^e_{jJ}V^e_{kK}, \\
    \ps{}{4}T^r_{IJKL} &= V^e_{iI}\ps{}{4}T_{ijkl}U_{jJ}V_{kK}V_{lL}, \\
    \ps{}{5}T^r_{IJKLM} &= V^e_{iI}\ps{}{5}T_{ijklm}U_{jJ}U_{kK}V_{lL}V_{mM},
\end{align*}
and where we explicitly use~$V^e$ to show that we use the part of~$\V$ corresponding to the concerned element.

\subsection{Reduced actuation force} \label{sec:reduced_actuation_force}

Finally, the reduced counter-part of the element-level actuation force~\eqref{eq:actuation_force} is, using Einstein notation,
\begin{equation*}\label{eq:reduced_actuation_force}
    f_L^r = \frac{k}{2}a V^e \big(\ps{}{1}B_L + \ps{}{2n}B_{Li}\eta_i + \ps{}{2\xi}B_{Li}\xi_{i}\big)
\end{equation*}
where
\begin{align*}
     \ps{}{1}B^e_{I} 
     & 
     = 2V^e_{nI}m_{V,i}H_{ij}G_{jn}
     \\
     \ps{}{2n}B^e_{IJ} 
     &
     = V^e_{nI}m_{V,i}(L_{1,ijk}G_{kl}V^e_{lJ}G_{jn}
     \\
     &+L_{1,ijk}G_{kn}G_{jm}V^e_{mJ})
     \\
     \ps{}{2\xi}B^e_{IJ} 
     &
     = V^e_{nI}m_{V,i}(L_{1,ijk}G_{kn}G_{jm} U^e_{mJ} 
     \\
     &+ 2L_{1,ijk}G_{kl}U^e_{lJ}G_{jn}).
\end{align*}
\FloatBarrier
\section{Optimization using PROM gradients}\label{sec:05_optimization_pipeline}

In this section, we present our optimization pipeline (\Cref{fig:optimization_pipeline}), which leverages the PROM developed in~\Cref{sec:04_PROM_section}. 
\begin{figure}[tb]
    \centering
    \newcommand{\xshift}{0.0}  

\begin{tikzpicture}[auto,node distance=0pt]

\tikzstyle{block} = [rectangle, draw, text width=12em, text centered, minimum height=3.0em]
\tikzstyle{block_if} = [rectangle, draw, text width=12em, minimum height=2.0em]
\tikzstyle{block_then} = [rectangle, draw, text width=10em, text centered, minimum height=2.5em]
\tikzstyle{arrow} = [thick,->,>=stealth]
\tikzstyle{reftext} = [text width=6em, align=right]

\def\dist{-1.4}
\node [block] (box1) {Upload nominal mesh, define shape variations};
\node [block, below of=box1,yshift=\dist cm] (box2) {Build PROM};
\node [block, below of=box2,yshift=\dist cm] (box3) {Solve EoMs and sensitivity};
\node [block, below of=box3,yshift=\dist cm] (box4) {Evaluate cost function\\ and its gradient};
\node [block, below of=box4,yshift=\dist cm] (box5) {Update parameters};
\node [block_if, below of=box5,yshift=-1.2 cm] (box6) {\textit{if parameters $<$ threshold}};
\node [block_then,below = -\the\pgflinewidth of box6.south west, anchor = north west] (box7) {Approximate EoMs solutions (sensitivity)};
\node [block_if, below= -\the\pgflinewidth of box7.south west, anchor = north west] (box8) {\textit{else}};
\node [block_then, below= -\the\pgflinewidth of box8.south west, anchor = north west] (box9) {New PROM/EoMs solutions needed};

\node[anchor=west] at ($(box1.east)+(\xshift,0)$) {\Cref{fig:shape_variations_procedure}};
\node[anchor=west] at ($(box2.east)+(\xshift,0)$) {\Cref{sec:04_PROM_section}};
\node[anchor=west, text width=2.5cm] at ($(box3.east)+(\xshift,0)$) {Eqns~\eqref{eq:red_EoMs},~\eqref{eq:def_sensitivity}};
\node[anchor=west, text width=2.5cm] at ($(box4.east)+(\xshift,0)$) {Eqns~\eqref{eq:cost_function},~\eqref{eq:cost_function_gradient}};
\node[anchor=west, text width=2.5cm] at ($(box5.east)+(\xshift,0)$) {Eqn~\eqref{eq:parameter_updates}};

\node[anchor=west, text width=2.5cm] at ($(box7.east)+(\xshift,0)$) {Eqn~\eqref{eq:approximation_using_sensitivity}};

\draw [arrow] (box1) -- (box2);
\draw [arrow] (box2) -- (box3);
\draw [arrow] (box3) -- (box4);
\draw [arrow] (box4) -- (box5);
\draw [arrow] (box5) -- (box6);
\draw [arrow, bend right=90] (box9.west) -- ++(-0.8cm,0) |- (box2.west);
\draw [arrow, bend right=90] (box7.west) -- ++(-0.4cm,0) |- (box4.west);

\end{tikzpicture}
    \caption{Optimization pipeline described in \Cref{sec:05_optimization_pipeline}. The algorithm runs until convergence.}
    \label{fig:optimization_pipeline}
\end{figure}
The pipeline starts with the upload of a nominal mesh and the definition of shape variations. The linear combination of these shape variations is defined by the parameter vector~$\xiv$ (see \Cref{fig:shape_variations_procedure} and \Cref{sec:PROM_ROB}). Then, a first PROM is built based on this nominal mesh. 

Based on the search parameters included in the vector~$\xiv$, the optimization objective is the maximization of the distance swam by the fish during a fixed time horizon. Specifically, the cost function to minimize is
\begin{equation}\label{eq:cost_function}
    L = \sum_{i=n}^N -\dvec_{\mathrm{swim}}\cdot \V \etav_i(\xiv),
\end{equation}
\noindent where~$\dvec_{\mathrm{swim}}$ is the swimming direction expressed at the assembly-level, and the indices $i=n, n+1,...,N$ correspond to discrete time steps for which the solutions of the EoMs are computed. Our PROM formulation allows to use gradient descent to find local minima of the cost function. In particular, the gradient of the cost function is given by
\begin{equation}\label{eq:cost_function_gradient}
   \nabla L = \sum_{i=n}^N -\dvec_{\mathrm{swim}}\cdot \V \sens_i,
\end{equation}
where
\begin{equation}\label{eq:def_sensitivity}
   \sens =  \der{\etav}{\xiv}
\end{equation}
is referred to as the \textit{sensitivity} and describes how the solution~$\etav$, computed for the nominal shape, changes for infinitesimal changes in the parameters~$\xiv$. It can be obtained by solving as a separate ordinary differential equation (ODE) or more efficiently directly when solving the EoMs as shown by \cite{bruls2008sensitivity}.  This step relies on the analytical expressions of the force derivatives, which can easily be obtained due to the tensorial structure of the reduced forces. In addition, we include barrier functions in the cost function~\eqref{eq:cost_function} to be able to add linear constraints to the parameter vector~$\xiv$.

The gradient of the cost function is then used to update the parameter vector~$\xiv$. Denoting this vector at a given optimization step $k$ as $\xiv_k$, the algorithm performs a gradient descent step with user-defined learning rate~$\gamma$,
\begin{equation}\label{eq:parameter_updates}
    \xiv_{k+1} = \xiv_{k} -\gamma \mathbf{W}_k \nabla L_k,
\end{equation}
where $\mathbf{W}_k$ is a diagonal weight matrix which is adapted during optimization to ensure convergence and constraint satisfaction. \MD{ More precisely, it decreases the learning rate along specific directions if constraints are active in these directions.} The computed sensitivity is also used to approximate the EoMs solutions in a neighborhood of the nominal solution~$\etav_0$ for the updated parameter vector~$\xiv_{k+1}$:
\begin{equation}\label{eq:approximation_using_sensitivity}
    \etav_{k+1} \approx \etav_0 + \sens\cdot \xiv_{k+1} .
\end{equation}
Because the approximation~\eqref{eq:approximation_using_sensitivity} is only valid in the vicinity of the nominal solution $\etav_0$, the pipeline checks if the updated parameters in~$\xiv_{k+1}$ are above a certain threshold. If this is the case, it builds a new PROM, considering the updated fish shape described by $\xiv_{k+1}$ as the new nominal shape. This algorithm runs until it reaches convergence, i.e., until the cost function does not change between successive optimization steps.

\section{Results and discussion}\label{sec:06_results}

To demonstrate the benefits of our approach, we present \MD{three} types of numerical experiments. Firstly, \Cref{sec:res_sim_setup} describes the chosen ROB and shows how the FOM, ROM and PROM relate in terms of computational costs and accuracy. Secondly, \Cref{sec:shape_optimization} applies the optimization pipeline developed in \Cref{sec:05_optimization_pipeline} to find optimal fish shapes for different numbers of search parameters. \MD{Finally, \Cref{sec:multi_obj_co_opti} extends the shape optimization pipeline to account for actuation signal optimization, enabling to tackle multi-objective optimization.} Our simulations are performed using the open-source FEM package ``YetAnotherFEcode'' for Matlab \citep{YetAnotherFEcode}, which we extended with the material presented above. The numerical case studies are run on a laptop equipped with an AMD Ryzen 7 PRO 6850U CPU with 32 GB of RAM, and the code is available online (\url{https://github.com/srl-ethz/AquaROM}).

\subsection{Simulation setup and model comparisons}\label{sec:res_sim_setup}

In this section, we report the simulation setup used for our case studies  \MD{and assess its accuracy}. \MD{The key elements of the simulation setup are: the nominal FEM structure and its properties (\Cref{sec:FEM_nominal_structure}), the ROB selection (\Cref{sec:res_ROB_selection}), and the solver parameters (\Cref{sec:simulation_parameters}). The simulation framework enables the fast solving of high-accuracy (P)ROMs (\Cref{sec:res_model_comparisons}).}


\subsubsection{FEM nominal structure}\label{sec:FEM_nominal_structure}

\begin{figure}[tb]
    \centering
    \includegraphics[width=\linewidth]{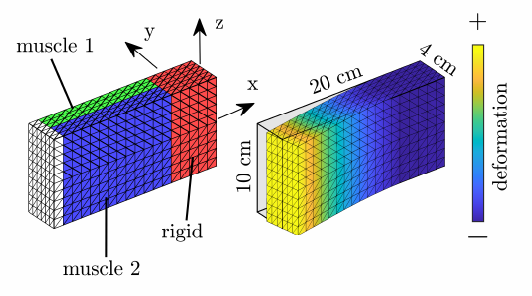}
    \caption{Nominal structure with a length of 20\si{cm}, a width of 4\si{cm}, and a height of 10\si{cm}. The left image shows the two muscles (green and blue), as well as the rigid part of the structure (red). The right image shows the first vibration mode (VM), which is an oscillation of the fish' tail.}
    \label{fig:A_muscles_placement_VM}
\end{figure}

Each of our numerical experiments starts with the nominal structure depicted in \Cref{fig:A_muscles_placement_VM}. It consists of a rectangular block equipped with two muscles. The elements of these muscles contract and extend according to the actuation forces described in \Cref{sec:actuation_force}. \MD{The head is rigid, following existing robotic designs~\citep[e.g.][]{katzschmann2018exploration,gravert_low-voltage_2024}.} 

\MD{The simulations are performed using a Saint Venant-Kirchhoff constitutive model and linear tetrahedra FEs, the numbers of which vary across the numerical experiments. The Saint Venant-Kirchoff model is computationally easy to compute but can lead to poor behavior for large strains. In the nominal case, the maximum principal strain is less than 5\% and localized in a few FEs in the outer layer of the structure, which justifies its use in our case studies. In particular, the tail's tip reach large deformations (in the order of 50\% of the body's width), but the maximum strain remains small. We discuss extensions to other constitutive models (e.g., co-rotated or Neo-Hookean) and larger strains as future work in~\Cref{sec:07_conclusion_future_work}.} 

\MD{In the simulation and optimization case studies, we focus on forward swimming behaviors:} the first part of the fish, composed of the rigid head, is constrained in the vertical and lateral directions so that it can only move in the forward or backward direction. \MD{The imposed motion constraints reflect the typical locomotion patterns observed in real fishes during steady forward swimming~\citep{shelton2014undulatory}. Focusing on forward swimming allows us to provide an intuitive illustration of the PROM-based optimization pipeline developed in \Cref{sec:05_optimization_pipeline}. While the underlying force reduction and optimization framework readily extend to more complex swimming tasks, analyzing such scenarios lies outside the scope of this paper, as they typically require more advanced control strategies~\citep[e.g.,][]{SoftCon,liu2022fishgym}.}

\subsubsection{ROB selection}\label{sec:res_ROB_selection}

The ROB used for our PROM consists of a rigid body mode along the forward direction ($x$-direction in \Cref{fig:A_muscles_placement_VM}), the first VM of the structure (right image of \Cref{fig:A_muscles_placement_VM}), its corresponding MDs, and the PSs associated with the shape parameters used in each different simulation. The selection of the shape variations, while not the focus of this first results section, is the same as the selection used in the shape optimization experiences described in \Cref{sec:shape_optimization} and \Cref{fig:all_shape_variations}. \MD{In particular, the shape variations used in the optimization cases SO1, SO2, and SO3, use 3, 5 and 8 shape variations represented in \Cref{fig:all_shape_variations}}. We also consider cases without any shape parameter, denoted as ROM (in comparison to PROM). 

\MD{We show that the ROM constructed with the selected ROB is accurate by comparing it to alternative ROB selections (\Cref{sec:appendix_rob_analysis}).} 

\subsubsection{\MD{Solver} parameters}\label{sec:simulation_parameters}

\MD{In order to solve the EoMs~\eqref{eq:EoMs_nonlinear} and~\eqref{eq:red_EoMs}, several numerical parameters need to be specified.}
The EoMs are solved using the Newmark-$\beta$ integration scheme~\citep{geradin2015mechanical}, with a time step of 0.02~\si{s} and a horizon of 2~\si{s}. \MD{The sensitivity~\eqref{eq:def_sensitivity} is computed without solving a separate ODE, but equivalently by using the Jacobian of the EoMs and computing the residual with respect to the optimization parameters, following~\cite{bruls2008sensitivity}}. The optimization horizon of the cost function~\eqref{eq:cost_function} is set to the last 0.1~\si{s} ($n=96$, $N=101$). 

We set the density of the fluid to $\rho=1000$~\si{kg/m^3}. The actuation signals applied to the muscle elements shown in \Cref{fig:A_muscles_placement_VM} are opposite sinusoids with a frequency of 1~\si{Hz} and an amplitude~$a=0.2$. The actuation stiffness $k$ varies between experiments, affecting the overall oscillation of the tail. We choose these simple-to-model signals because the focus of this work is shape optimization, and discuss limitations and future developments for actuation optimization in \Cref{sec:07_conclusion_future_work}. \MD{First results in that direction are presented in \Cref{sec:multi_obj_co_opti}.}

\subsubsection{Model comparisons}\label{sec:res_model_comparisons}
\begin{figure}[tb]
    \centering
    \includegraphics[width=\linewidth]{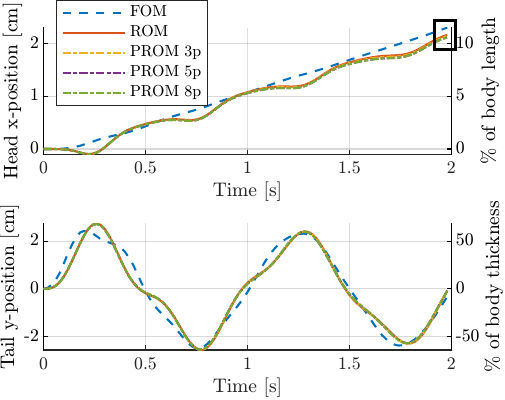}
    \caption{Comparison of the solutions obtained by simulating the FOM, ROM, and PROM for multiple share variation parameters (3,5, and 8 parameters), for a mesh discretized in 8086 FEs and an actuation amplitude of $k=8.5\cdot 10^4$. The upper panel shows the swimming distance achieved by the fish (head position along the $x$-direction), and the lower panel shows the oscillation of the tail ($y$-position) across time. The error between the FOM and (P)ROMs after 2~\si{s} (rectangle in the upper panel) is reported in \Cref{fig:A_FOM_ROM_comparison}.}
    \label{fig:A_accuracy}
\end{figure}

To compare the FOM, ROM, and PROM formulations, we simulate these models for five different mesh discretizations (between 1272 and 24822 FEs), and multiple actuation stiffnesses $k$ (\MD{between $1.9\cdot10^4$ and $14.5\cdot 10^4$}), resulting in different amplitudes of the tail oscillation. For each combination of mesh discretization and actuation stiffness, we compare the solution obtained by simulating the ROM and PROM to the solution obtained by simulating the FOM. In particular, we focus on the head position in the swimming direction ($x$-position) and the tail position in the oscillation direction ($y$-position), as shown in \Cref{fig:A_accuracy} for the example of a mesh discretization of 8086 elements and an actuation stiffness of \MD{$k=8.5\cdot 10^4$}. Because the ROB used by the different PROMs is similar to the one used by the ROM (with the additional basis vectors contained in the matrix $\Xi$), their simulation results for the nominal shape are the same. Compared to the FOM, all (P)ROMs yield a smaller swimming distance (\Cref{fig:A_accuracy}, upper panel), due to their inability to fully capture the full oscillation of the tail (\Cref{fig:A_accuracy}, lower panel). Specifically, the relative error in the swimming distance between the FOM and the (P)ROM is \MD{\mbox{-6\%}} percent after 2~\si{s} (see rectangle in \Cref{fig:A_accuracy}).

\begin{figure}[tb]
    \centering
    \includegraphics[width=\linewidth]{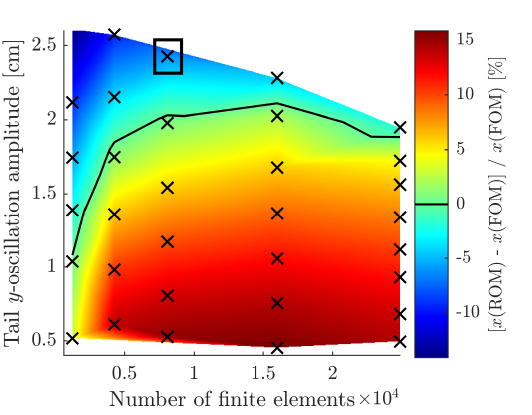}
    \caption{Relative error in the swimming distance after $t=2$\si{s} between the FOM and the ROM for multiple discretizations and oscillation's amplitudes of the nominal shape (see \Cref{fig:A_muscles_placement_VM}). The crosses indicate simulations results, and the heat map is obtained using a linear interpolation between these results. Simulations of the FOM performed in the white area in the top right corner (i.e., for large number of FEs and large oscillations) did not converge. The rectangle corresponds to the relative error measured in \Cref{fig:A_accuracy}.}
    \label{fig:A_FOM_ROM_comparison}
\end{figure}

\MD{To compare the models more globally, we repeat the analysis of \Cref{fig:A_accuracy} for multiple combinations of mesh discretization and oscillation amplitudes. The results are shown in \Cref{fig:A_FOM_ROM_comparison}.} As an example, the relative error reported for the case presented in \Cref{fig:A_accuracy} is highlighted by a rectangle in both figures. The relative error remains small for most combinations (between \MD{-10\% and +15\%}), \MD{and in particular for an amplitude of around 2\si{cm}, where we operate the system in \Cref{sec:shape_optimization}}. \MD{While the relative error grows for small oscillation amplitudes, the absolute error decreases (\Cref{fig:FOM_PROM_abs_error_profile}).} \MD{The accuracy of the (P)ROMs is strenghten by the fact that other choices of muscle placements also lead to a meaningful error profile (\Cref{fig:FOM_PROM_alternative_error_profile}), and the solver operates in a regime of convergence under the chosen time step (\Cref{sec:appendix_simulation_convergence})}

\begin{figure}[tb]
    \centering
    \includegraphics[width=0.95\linewidth]{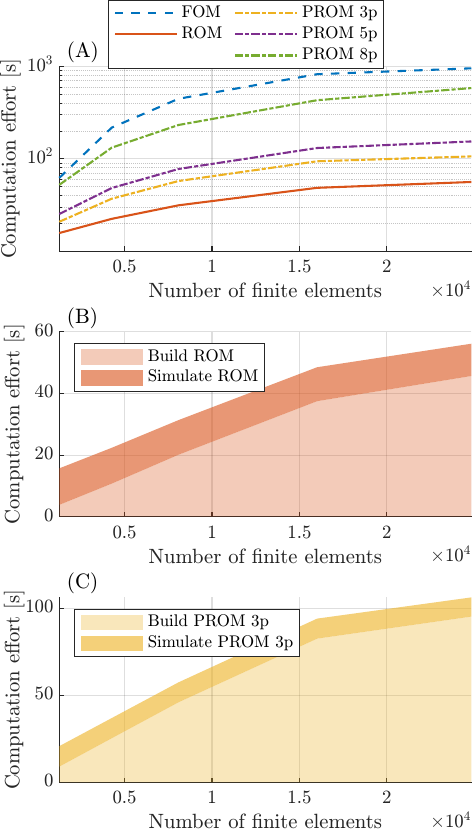}
    \caption{Computation effort as a function of the number of finite elements, for different models. Panel (A) shows the total computation effort for the FOM, ROM and 3 PROMs with resepectively 3, 5, and 8 parameters. Panels (B) and (C) decompose the cost for the ROM and the PROM with 3 parameters into the time needed to build the model and the time needed to simulate it.}
    \label{fig:A_comp_time}
\end{figure}

While simulation results of (P)ROMs are similar to those obtained with the FOM, they require significantly less computation efforts. Panel (A) of \Cref{fig:A_comp_time} shows the simulation time required to solve the EoMs for the different models and mesh discretizations, averaged over the different oscillation amplitudes of \Cref{fig:A_FOM_ROM_comparison}. As can be expected, the simulation time increases with the number of FEs. Comparing the different models, the time required by the ROM is approximately one order of magnitude smaller compared to the time required by the FOM. The simulation of the different PROMs requires more computational effort than the ROM because of their enhanced ROB. This computational effort increases with the number of parameters used to describe the shape variations. 

Panels (B) and (C) of \Cref{fig:A_comp_time} split the computation effort of the ROM and the PROM with 3 parameters into two parts: the time needed to build the model (i.e., to construct the reduced order tensors) and the time needed to solve the EoMs. For both cases, the total computation effort is driven by the model building, which increases with the number of FEs, while the time needed to simulate the system, i.e., to solve the EoMs, is constant for the different discretization. This pattern is explained by the fact that the reduced order tensors are constructed by considering the contribution of each FE, but the dimensionality of the EoMs~\eqref{eq:red_EoMs} is independent of the mesh discretization. While the PROMs require more computational resources compared to the ROM, they allow the computation of analytical sensitivities, which are leveraged in the optimization pipeline introduced in \Cref{sec:05_optimization_pipeline}, whose performance is analyzed next.

\subsection{Shape optimization}\label{sec:shape_optimization}

\begin{figure*}[tb]
    \centering
    \includegraphics[width=\linewidth]{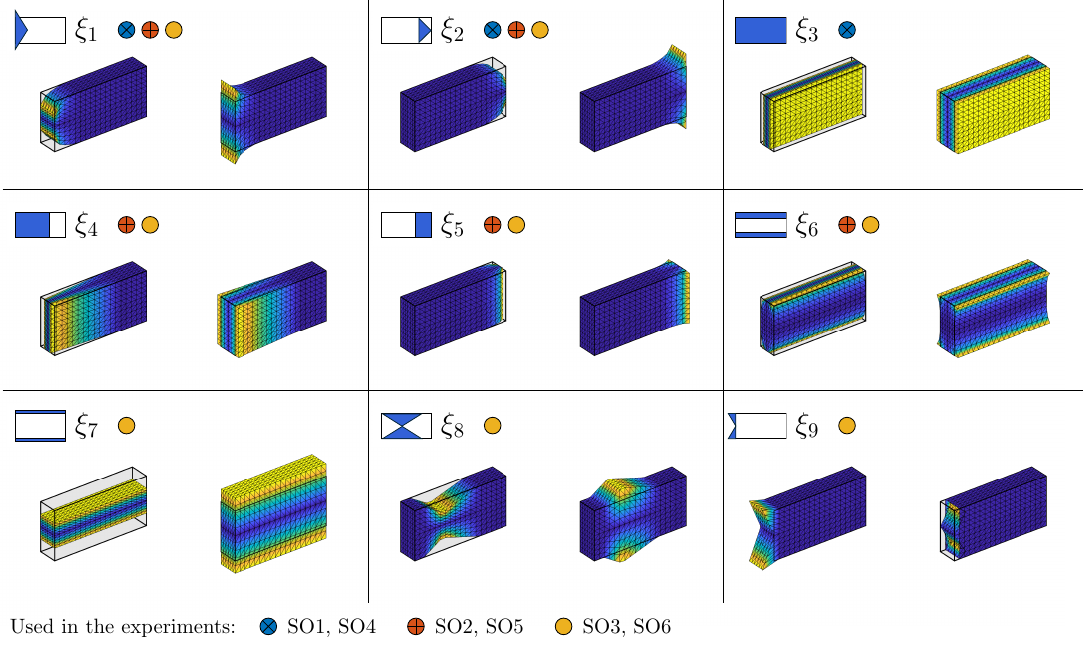}
    \caption{Shape variations used in the different numerical experiments. Each panel corresponds to a different shape variation obtained using the method  described in \Cref{sec:PROM_ROB} and \Cref{fig:shape_variations_procedure}, i.e., a single column of $\U$. In each panel, the $i$-th column of $\U$ is multiplied by $\xi_i=0.5$ (left shape), and $\xi_i=-0.5$ (right shape). The numerical experiments combine the shape variations according to the round symbols to build the matrix $\U$, e.g., SO1 uses the shape variations $i=1,2,3$ to obtain $\U\in\mathbb{R}^{n\times 3}$.}
    \label{fig:all_shape_variations}
\end{figure*}

We now turn to the shape optimization (SO) of soft swimmers, using the pipeline shown in \Cref{fig:optimization_pipeline}. Starting from a uniform 3-dimensional block (\Cref{fig:A_muscles_placement_VM}), our goal is to find an optimal shape that maximizes the distance traveled in~$x-$direction by the fish within a 2 seconds time horizon. We present 6 \MD{primary} numerical experiments, summarized in \Cref{tab:results_summary}.  These experiments differ in the discretization of the mesh and in the number of optimization parameters. The optimization parameters correspond to different shape variations, which can be linearly combined to find the final optimal shape (overview in \Cref{fig:all_shape_variations}). \MD{After presenting the results of these 6 experiments (Sections~\ref{sec:SO1} to~\ref{sec:SO46}), we discuss robustness checks and additional results in \Cref{sec:robustness_checks}. Finally, \Cref{sec:discussoin_comparison} discusses comparisons to other non-gradient based optimization strategies.}
\begin{table*}[tb]
\caption{Summary of the numerical case studies. The column ``\# Build \& EoMs'' reports the number of times the PROM was built and the EoMs solved, according to the algorithm presented in~\Cref{fig:optimization_pipeline}. Accordingly, the far-right column reports the average time needed to build the PROM and solve the EoMs once, i.e., for one outer loop iteration of the algorithm.\label{tab:results_summary}}
\footnotesize
\begin{center}
\renewcommand{\arraystretch}{1.3}
\rowcolors{2}{white}{gray!25}
\begin{tabular}{c|ccccc|cc}
\toprule
Case & \# Parameters & \# FEs & \# DoFs FOM & \# DoFs PROM  & \# Build \& EoMs & Time [min.] & Time/(Build \& EoMs) [min.] \\ \midrule 
\SOOneSymbol$\,$    SO1 & 3 & 8086  & 5181  & 6 & 6 & 5.74  & 0.96  \\
\SOTwoSymbol$\,$    SO2 & 5 & 8086  & 5181  & 8 & 6 & 7.86  & 1.31  \\
\SOThreeSymbol$\,$  SO3 & 8 & 8086  & 5181  & 11 & 6 & 22.34 & 3.72  \\
\SOOneSymbol$\,$    SO4 & 3 & 16009 & 9714  & 6 & 5 & 14.11 & 2.82  \\ 
\SOTwoSymbol$\,$    SO5 & 5 & 16009 & 9714  & 8 & 6 & 21.90 & 3.65  \\
\SOThreeSymbol$\,$  SO6 & 8 & 16009 & 9714  & 11 & 6 & 68.59 & 11.43 \\\bottomrule
\end{tabular}

\end{center}
\end{table*}

\subsubsection{Experiment SO1}\label{sec:SO1} \SOOneSymbol The search space of the first experiment SO1 is composed of the 3 shape variations presented in the upper row of \Cref{fig:all_shape_variations}. Therefore, the parameter vector~$\xiv$ contains 3 parameters. We assign the following constraints to the parameters:~$-0.5 \leq \xi_1,\xi_2\leq0.5$ and~$-0.15\leq \xi_3\leq 0.15$. These constraints are summarized in \Cref{tab:constraint_summary}.

\begin{table}[tb]
\footnotesize
\caption{Lower and upper bound on the search parameters for the different shape optimization experiments.\label{tab:constraint_summary}
}
\begin{center}
\renewcommand{\arraystretch}{1.3}
\rowcolors{2}{white}{gray!25}
\begin{tabular}{c|ccc}
\toprule
Case & \SOOneSymbol$\,$    SO1, SO4 & \SOTwoSymbol$\,$    SO2, SO5 & \SOThreeSymbol$\,$  SO3, SO6 \\ \midrule 
$\xi_1$ & $-0.5, 0.5$   & $-0.5, 0.5$   &  $-0.5, 0.5$  \\
$\xi_2$ & $-0.5, 0.5$   & $-0.5, 0.5$   &  $-0.5, 0.5$  \\
$\xi_3$ & $-0.15, 0.15$   & not used      & not used      \\
$\xi_4$ & not used      & $-0.4, 0.4$   &  $-0.3, 0.3$  \\
$\xi_5$ & not used      & $-0.5, 0.5$   &  $-0.4, 0.4$  \\
$\xi_6$ & not used      & $-0.5, 0.5$   &  $-0.4, 0.4$  \\
$\xi_7$ & not used      & not used      &  $-0.25, 0.25$  \\
$\xi_8$ & not used      & not used      &  $-0.25, 0.25$  \\
$\xi_9$ & not used      & not used      &  $-0.01, 0.2$  \\\bottomrule
\end{tabular}
\end{center}
\end{table}

\begin{figure}[tb]
    \centering
    \includegraphics[width=0.95\linewidth]{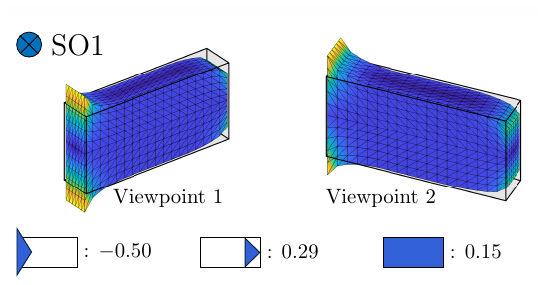}
    \caption{Optimal shape obtained for SO1, presented from two different viewpoints. The symbols and their values correspond to the optimal parameters obtained for each shape variation of the search space (\Cref{fig:all_shape_variations}).}
    \label{fig:SO1_opt_shape}
\end{figure}

The obtained optimal shape is depicted in \Cref{fig:SO1_opt_shape}. As could be intuitively expected, this simple example shows the creation of a thinner fish with a tail and a head, therefore maximizing its propulsion. The optimal fish traveled a distance 3.8 larger than the nominal fish (\Cref{fig:SO_dist_evolution}).

This numerical case study is fast to solve, as reported in \Cref{tab:results_summary}. It took 5.74~\si{min} to converge and required 6 rebuilds of the PROM and resolving of the EoMs (see \Cref{fig:optimization_pipeline}), leading to an average of 0.96~\si{min} per build and EoMs solve.

\subsubsection{Experiment SO2}\label{sec:SO2} 
\begin{figure}[tb]
    \centering
    \includegraphics[width=0.95\linewidth]{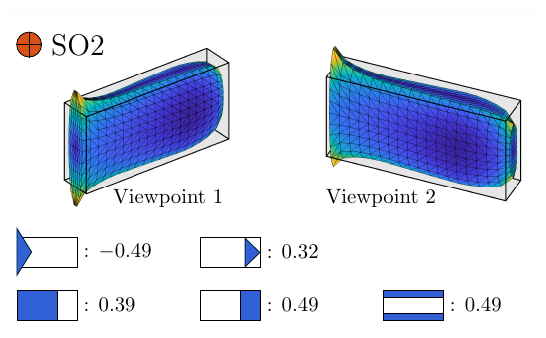}
    \caption{Optimal shape obtained for SO2, presented from two different viewpoints. The symbols and their values correspond to the optimal parameters obtained for each shape variation of the search space (\Cref{fig:all_shape_variations}).}
    \label{fig:SO2_opt_shape}
\end{figure}

\SOTwoSymbol In this experiment, we augment the search space of SO1 replacing the shape~$\xi_3$ by the three shape variations in the second row of \Cref{fig:all_shape_variations}, for a total of 5 search parameters. Similarly to SO1, the optimal shape (\Cref{fig:SO2_opt_shape}) allows to outperform the nominal shape by a factor of 5.60 (\Cref{fig:SO_dist_evolution}). Because of the larger search space, SO2 requires slightly more computational effort than SO1 and converges in 7.86~\si{min} (1.31~\si{min} per build and EOMs solve).

\subsubsection{Experiment SO3}\label{sec:SO3} 
\begin{figure}[tb]
    \centering
    \includegraphics[width=0.95\linewidth]{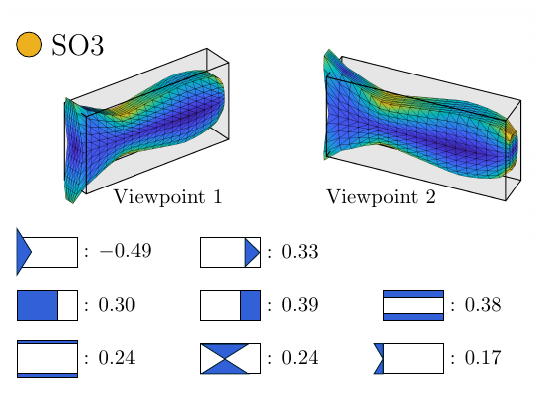}
    \caption{Optimal shape obtained for SO3, presented from two different viewpoints. The symbols and their values correspond to the optimal parameters obtained for each shape variation of the search space (\Cref{fig:all_shape_variations}).}
    \label{fig:SO3_opt_shape}
\end{figure}

\begin{figure}[tb]
    \centering
    \includegraphics[width=0.95\linewidth]{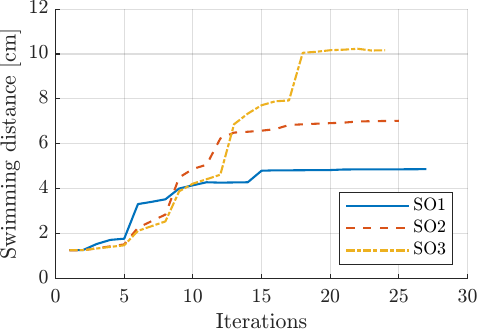}
    \caption{Swimming distance across optimization iterations. After convergence, the optimal shapes obtained in the experiments SO1, SO2, and SO3 clearly outperform the nominal shape. The optimal shape obtained with SO3 allows for larger swimming distance compared to SO2 and SO1, although the difference is relatively small.}
    \label{fig:SO_dist_evolution}
\end{figure}

\SOThreeSymbol
The numerical experiment SO3, which includes 8 parameters as shown in \Cref{fig:all_shape_variations}, results in the optimal shape \Cref{fig:SO3_opt_shape}. Compared to SO1 and SO2, the optimal shape is characterized by a more slender forebody and an enlarged tail, given the smaller body size. This optimal shape allows us to outperform the nominal shape by a factor 8 and the optimal shapes obtained in SO1 and SO2 as shown in \Cref{fig:SO_dist_evolution}. In terms of computational effort, SO3 converges in 12.6~\si{min}, which is fast but slower than SO1 (factor of 3.2) and SO2 (factor of 2.3) due to the larger number of search parameters.

\subsubsection{Experiments SO4 to SO6}\label{sec:SO46} To assess the properties of our method for finer meshes, we re-run the experiments SO1 to SO3 using a finer FE mesh. The main characteristics of these case studies are reported in~\Cref{tab:results_summary}. The optimal shapes are identical to those of SO1-SO3 within the numerical accuracy limits of our pipeline and therefore are not explicitly shown. Because of the finer mesh discretization, the optimization requires more time to run (approximately a factor 3 compared to SO1-SO3), an increase which is mainly driven by the additional time required to build the PROMs (see \Cref{fig:A_comp_time}).

\subsubsection{Robustness checks}\label{sec:robustness_checks}
\MD{We run multiple robustness checks that we report in \Cref{sec:appendix_additional_optimization_results}. The key insights from these additional results are the following: 
\begin{itemize}
    \item The obtained optimal parameters are likely to represent global optima. Different parameter initializations leads to the same optimal values (\Cref{sec:appendix_opt_initialization}).
    \item The solution approximations based on the PROM sensitivities~\eqref{eq:approximation_using_sensitivity} used in the optimization pipeline represent a trade-off. Using the gradient computed based on these approximations allows to avoid rebuilding a PROM, but it might not be accurate. In the cases we consider, the descent direction remains precise in the interval between two PROM builds (\Cref{sec:appendix_PROM_approx}).  
    \item The quality of the design search space ultimately depends on the expressiveness of not only the single shape variations, but also their linear combinations. The optimization pipeline allows practitioners to rapidly test different designs of a specific morphological feature (e.g., the fish tail) while respecting manufacturing constraints (\Cref{sec:appendix_alternative_shapes}).
\end{itemize}}

\subsubsection{Discussion and comparison}\label{sec:discussoin_comparison}

\MD{The results presented above show that the PROM-based optimization reliably finds optima in the search space, and is especially fast for up to approximately 5 parameters. The PROM sensitivities allow to compute gradient of the cost function and therefore efficiently search the parameter space.}

\MD{The efficiency of the PROM-based approach is particularly clear when compared to the potential for a FOM- or ROM-based algorithms. The FOM is significantly more costly to run (\Cref{fig:A_comp_time}) and does not provide analytical sensitivities with respect to the shape parameters. An alternative would be to consider the ROM and perform a grid search without gradient. However, this is way less efficient than the PROM-based method: \Cref{tab:comparison_grid_search} shows that for a similar granularity of the search space as the rebuild threshold of the PROM-based algorithm, a ROM grid search would suffer from the curse of dimensionality with few parameters already. Even when solving for the corners of the search hyperbox, a ROM-based model is slower in most cases. These results confirm the appropriateness of the proposed optimization scheme.}

\begin{table}[tb]
\footnotesize
\caption{Computational effort estimation for a ROM-based grid search. The time per ROM solve is taken from \Cref{sec:res_model_comparisons}.\label{tab:comparison_grid_search}
}
\begin{center}
\renewcommand{\arraystretch}{1.3}
\rowcolors{2}{white}{gray!25}
\begin{tabular}{c|ccc}
\toprule
Case & \SOOneSymbol$\,$    SO1 & \SOTwoSymbol$\,$    SO2& \SOThreeSymbol$\,$  SO3 \\ \midrule 
Number of parameters        & 3     & 5     & 8\\
Number of corners           & 8     & 32    & 256    \\
Number of combinations      & 147   & 14406 & 367500\\
Single ROM solve time [s]   & 31    & 31    & 31\\
Time for corners only [min] & 4.1   & 16.5  & 132 \\
Time for all combi. [h]     & 1.26  & 124   & 3165  \\
Time for PROM [min]         & 5.7   & 7.86  & 22.3 \\\bottomrule
\end{tabular}
\end{center}
\end{table}

\subsection{Multi-objective co-optimization}\label{sec:multi_obj_co_opti}

\MD{We illustrate how the shape optimization pipeline can be extended to a multi-objective, co-optimization framework by focusing on two competing objectives: a large swimming distance and a low energy use. To describe the energy use, we parametrize the sinusoidal actuation signal by its amplitude $a$ and consider the following quadratic actuation energy:
\begin{equation}\label{eq:cost_energy}
    L_{\mathrm{energy}} = \alpha a^2,
\end{equation}
where $\alpha$ is a constant. The resulting multi-objective cost function is
\begin{equation}
    L = w_1 L_{\mathrm{distance}} + w_2 L_{\mathrm{energy}},
\end{equation}
where $L_{\mathrm{distance}}$ is the cost from the swimming distance~\eqref{eq:cost_function}, and $w_1, w_2$ are weights such that $w_1+w_2=1$.}

\MD{The computation of the analytical sensitivities~\eqref{eq:def_sensitivity} needed to perform co-optimization of the shape and actuation signal is  straightforward, as the actuation parameter (i.e., the signal amplitude) affects the EoMs exclusively through the actuation force~\eqref{eq:actuation_force}. Concretely, this implies that the partial derivatives of all other forces with respect to the actuation parameter vanish. In contrast, the partial derivatives of the actuation force with respect to the shape parameters must be augmented by its derivative with respect to the actuation parameter. The same reasoning extends to more complex actuation parametrizations. }

\begin{figure}[tb]
    \centering
    \includegraphics[width=\linewidth]{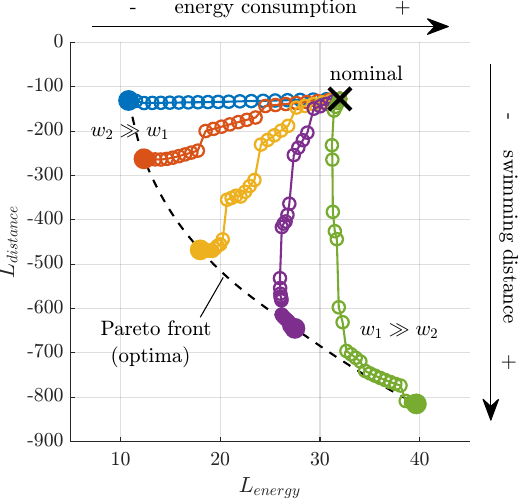}
    \caption{Multi-objective co-optimization results. Depending on the weights $w_1$ and $w_2$, different trade-off between energy consumption and swimming distance are achieved. Filled circles represent optima, while empty circles represent parameters discovered during the optimization.}
    \label{fig:co_optimization}
\end{figure}

\MD{We illustrate the multi-objective co-optimization in \Cref{fig:co_optimization}, where we use the shape variations of SO2 (total of 6 parameters) and vary the weights $w_1$ and $w_2$. When $w_2\gg w_1$, solutions with smaller actuation amplitudes are found, thereby reducing energy consumption. Conversely, when $w_1\gg w_2$, the optimization prioritizes achieving greater swimming distance. The set of optimal trade-offs forms a Pareto front shown in \Cref{fig:co_optimization}.}

\MD{While the actuation amplitude varies as expected at the different Pareto optima, the optimal fish morphology is barely affected. This is due to the fact that the energy cost $L_{\mathrm{energy}}$ is not a function of the fish shape, as the energy per muscle element is computed based on the nominal shape and not updated during optimization (\Cref{sec:actuation_force}). The only parameter explicitly coupling the two objectives is the actuation amplitude.}

\MD{This simple example illustrates that the shape optimization pipeline developed in this work can accomodate the co-optimization of fish morphologies and actuation signals. A comprehensive treatment of optimal control strategies lies beyond the scope of the present study but present an exciting extension for future work, which we discuss next.}

\section{Conclusion and future work}\label{sec:07_conclusion_future_work}

In this paper, we presented a novel shape optimization pipeline for reduced order models of mechanical FEM structures subject to nonlinear forces. \MD{To showcase the capabilities of this approach,} we applied \MD{it} to the task of finding optimal shapes of soft robotic fishes, which undergo large deformations and experience internal and external (hydrodynamic) nonlinear forces. We first validated the developed parametric reduced order models (PROMs) against full order models (FOMs) in terms of accuracy and computational effort. We then leveraged these accurate and computationally efficient PROMs, and in particular their corresponding parameter sensitivities, to perform gradient-based optimization. The resulting optimal shapes outperform the nominal shape by factors up to \MD{8} and are obtained rapidly (within a few minutes, see \Cref{tab:results_summary}), making \MD{the presented methodology well suited for fast, physics-based design optimization.}

The optimization pipeline developed in this paper shows, however, a few limitations. Although the accuracy of the PROMs was validated against FOMs, we did not compare our results against real-world robotic systems or more advanced hydrodynamic models. Instead, we relied on previous works that employed or assessed the Lighthill's hydrodynamic model~\citep{wang2011dynamic,li2014modeling, ma_diffaqua_2021, eloy2024flow}. Validating the optimal fish shapes against physical robotic prototypes is a promising direction for future work, which would allows assessing existing sim-to-real gaps~\citep{dubied2022sim}. \MD{Such comparisons would also clarify whether extending the reduction of internal forces used by \cite{Marconi2021AExpansion} to more advanced formulations -- such as co-rotational or Neo-Hookean constitutive models -- would meaningfully improve the fidelity of the PROM solutions.}

Importantly, our work advances existing methods using PROMs for optimization \citep{frohlich2019geometric} by incorporating nonlinear forces into the optimization pipeline. Future work should also consider the use \MD{of this general PROM-based optimization method} to other scenarios, and in particular to standard scenarios for which optimization can easily be performed using off-the-shelf optimizers, allowing for a thorough comparison. Finally, \MD{a more detailed treatement of richer actuation signals within co-optimization settings (\Cref{sec:multi_obj_co_opti}) could enable more practically relevant applications.}

Altogether, the optimization method presented in this paper shows how nonlinear PROMs can be used for optimization tasks in a computationally efficient way, unlocking new opportunities for more efficient robotic design and control.


\bibliographystyle{SageH}
\bibliography{references.bib}

\appendix
\section{Actuation force: matrices and tensors}\label{A_actuation_force}

In this section, we detail the expressions of the mathematical objects used to construct the actuation forces \eqref{eq:actuation_force} at the element-level (for 3D settings).

\begingroup
    \allowdisplaybreaks
    \begin{align}
         \I_V   &= [1, 1, 1, 0, 0, 0]^\top
         \\
         \Hm    &=\begin{bmatrix}
             1  &   0   &   0   &   0   &   0   &   0   &   0   &   0   &   0   \\
             0  &   0   &   0   &   0  &    1   &   0   &   0   &   0   &   0   \\
             0  &   0   &   0   &   0   &   0   &   0   &   0   &   0   &   1   \\
             0  &   1   &   0   &   1   &   0   &   0   &   0   &   0   &   0   \\
             0  &   0   &   1   &   0   &   0   &   0   &   1   &   0   &   0   \\
             0  &   0   &   0   &   0   &   0   &   1   &   0   &   1   &   0   \\
         \end{bmatrix}  
         \\\scriptsize
         \A_1(\thetav)  &= \scriptsize\begin{bmatrix}
             u_{,x} &0      &0      &v_{,x} &0      &0      &w_{,x} &0      &0      \\
             0      &u_{,y} &0      &0      &v_{,y} &0      &0      &w_{,y} &0      \\
             0      &0      &u_{,z} &0      &0      &v_{,z} &0      &0      &w_{,z} \\
             u_{,y} &u_{,x} &0      &v_{,y} &v_{,x} &0      &w_{,y} &w_{,x} &0      \\
             u_{,z} &0      &u_{,x} &v_{,z} &0      &v_{,x} &w_{,z} &0      &w_{,x} \\
             0      &u_{,z} &u_{,y} &0      &v_{,z} &v_{,y} &0      &w_{,z} &w_{,y}  \\             
         \end{bmatrix}
         \\\scriptsize
    \end{align}
\endgroup

$\Lm_1$ is a sparse third order tensor with elements $L^{(1)}_{ijk}$. \Cref{tab:L1_3D} shows the non-zero element of this tensor.

\begin{table}[htb]
    \centering
    \caption{Non-zero elements of the $\Lm_1$ third order tensor (3D case).
    \label{tab:L1_3D}}
    \begin{tabular}{ccccccccc}
       \toprule
        $L^{(1)}_{111}=1$ & $L^{(1)}_{144}=1$ & $L^{(1)}_{177}=1$ & $L^{(1)}_{222}=1$ & $L^{(1)}_{255}=1$ 
        \\[5pt]
        $L^{(1)}_{288}=1$ & $L^{(1)}_{333}=1$ & $L^{(1)}_{366}=1$ & $L^{(1)}_{399}=1$ & $L^{(1)}_{412}=1$
        \\[5pt]
        $L^{(1)}_{421}=1$ & $L^{(1)}_{445}=1$ & $L^{(1)}_{454}=1$ & $L^{(1)}_{478}=1$ & $L^{(1)}_{487}=1$ 
        \\[5pt]
        $L^{(1)}_{513}=1$ & $L^{(1)}_{531}=1$ & $L^{(1)}_{546}=1$ & $L^{(1)}_{564}=1$ & $L^{(1)}_{579}=1$
        \\[5pt]
        $L^{(1)}_{597}=1$ & $L^{(1)}_{623}=1$ & $L^{(1)}_{632}=1$ & $L^{(1)}_{656}=1$ & $L^{(1)}_{665}=1$
        \\[5pt]
        $L^{(1)}_{689}=1$ & $L^{(1)}_{698}=1$
        \\[2pt]
         \bottomrule
    \end{tabular}
    \vspace{2pt}
    
\end{table}
\section{Reactive force: matrices and tensors}\label{A_reactive_force}

This section details the tensors present in \eqref{eq:reduced_spine_force}. The reduced spine force at the element level is composed of three terms:

\begin{equation*}
    f_I^r = \textcolor{black}{\ps{0}{}f_I^r} + \textcolor{black}{\ps{1}{}f_I^r} + \ps{2}{}f_I^r
\end{equation*}
with 
{\color{black}
\begin{alignat*}{3}
    \ps{0}{}f^r_I 
    &= \ps{x x}{2}T^{}_{Ij}\etaddnv_j 
    &&
    + \ps{x U}{3}T^{}_{Ijk}\etaddnv_j \xi_k 
    & &
    + \ps{x V}{3}T^{}_{Ijk}\etaddnv_j \eta_k 
    \nonumber
    \\
    &
    + \ps{U x}{3}T^{}_{Ijk}\etaddnv_j\xi_k z_0
    &&
    + \ps{UU}{4}T^{}_{Ijkl}\etaddnv_j\xi_k\xi_l  z_0
    & &
    + \ps{U V}{4}T^{}_{Ijkl}\etaddnv_j\xi_k\eta_l z_0
    \nonumber
    \\
    & 
    + \ps{V x}{3}T^{}_{Ijk}\etaddnv_j\eta_k 
    &&
    + \ps{VU }{4}T^{}_{Ijkl}\etaddnv_j\eta_k\xi_l
    & &
    + \ps{V V}{4}T^{}_{Ijkl}\etaddnv_j\eta_k\eta_l 
    \nonumber
    \\
    &
    + \ps{V x}{3}T^{}_{Ijk}\etadnv_j\etadnv_k 
    &&
    + \ps{VU}{4}T^{}_{Ijkl}\etadnv_j\etadnv_k\xi_l
    & &
    + \ps{V V}{4}T^{}_{Ijkl}\etadnv_j\etadnv_k\eta_l
    \nonumber
    \\
    &
    + \ps{x V}{3}T^{}_{Ijk}\etadnv_j\etadnv_k
    &&
    + \ps{U V}{4}T^{}_{Ijkl}\etadnv_j\xi_k\etadnv_l
    & &
    + \ps{V V}{4}T^{}_{Ijkl}\etadnv_j\eta_k \etadnv_l,
\end{alignat*}
}
{\color{black}
\begin{alignat*}{3}
    \ps{1}{}f^r_I 
    &= \ps{x x}{3}T^{}_{Ijk}\etaddnv_j \xi_k
    &&
    + \ps{x U}{4}T^{}_{Ijkl}\etaddnv_j \xi_k \xi_l
    \nonumber
    \\
    &
    + \ps{x V}{4}T^{}_{Ijkl}\etaddnv_j \eta_k \xi_l
    &&
    + \ps{U x}{4}T^{}_{Ijkl}\etaddnv_j\xi_k  \xi_l
    \nonumber
    \\
    &
    + \ps{UU}{5}T^{}_{Ijklm}\etaddnv_j\xi_k\xi_l \xi_m
    &&
    + \ps{U V}{5}T^{}_{Ijklm}\etaddnv_j\xi_k\eta_l \xi_m
    \nonumber
    \\
    & 
    + \ps{V x}{4}T^{}_{Ijkl}\etaddnv_j\eta_k \xi_l
    &&
    + \ps{VU }{5}T^{}_{Ijklm}\etaddnv_j\eta_k\xi_l\xi_m
    \nonumber
    \\
    &
    + \ps{V V}{4}T^{}_{Ijklm}\etaddnv_j\eta_k\eta_l \xi_m
    &&
    + \ps{V x}{4}T^{}_{Ijkl}\etadnv_j\etadnv_k \xi_l
    \nonumber
    \\
    &
    + \ps{VU}{5}T^{}_{Ijklm}\etadnv_j\etadnv_k\xi_l\xi_m
    &&
    + \ps{V V}{5}T^{}_{Ijklm}\etadnv_j\etadnv_k\eta_l\xi_m
    \nonumber
    \\
    &
    + \ps{x V}{4}T^{}_{Ijkl}\etadnv_j\etadnv_k\xi_l
    &&
    + \ps{U V}{5}T^{}_{Ijklm}\etadnv_j\xi_k\etadnv_l\xi_m
    \nonumber
    \\
    &
    + \ps{V V}{5}T^{}_{Ijklm}\etadnv_j\eta_k \etadnv_l\xi_m,
\end{alignat*}
}
\begin{alignat*}{2}
    \ps{2}{}f^r_I 
    &= \ps{x x}{4}T^{}_{Ijkl}\etaddnv_j  \xi_k \xi_l
    &&
    + \ps{x U}{5}T^{}_{Ijklm}\etaddnv_j \xi_k \xi_l\xi_m
    \nonumber
    \\
    &
    + \ps{x V}{5}T^{}_{Ijklm}\etaddnv_j \eta_k \xi_l\xi_m
    &&
    + \ps{U x}{5}T^{}_{Ijklm}\etaddnv_j\xi_k \xi_l\xi_m
    \nonumber
    \\
    &
    + \ps{UU}{6}T^{}_{Ijklmn}\etaddnv_j\xi_k\xi_l \xi_m\xi_n
    &&
    + \ps{U V}{6}T^{}_{Ijklmn}\etaddnv_j\xi_k\eta_l\xi_m\xi_n
    \nonumber
    \\
    & 
    + \ps{V x}{5}T^{}_{Ijklm}\etaddnv_j\eta_k \xi_l\xi_m
    &&
    + \ps{VU }{6}T^{}_{Ijklmn}\etaddnv_j\eta_k\xi_l\xi_m\xi_n
    \nonumber
    \\
    &
    + \ps{V V}{6}T^{}_{Ijklmn}\etaddnv_j\eta_k\eta_l \xi_m\xi_n
    &&
    + \ps{V x}{5}T^{}_{Ijklm}\etadnv_j\etadnv_k \xi_l\xi_m
    \nonumber
    \\
    &
    + \ps{VU}{6}T^{}_{Ijklmn}\etadnv_j\etadnv_k\xi_l\xi_m\xi_n
    &&
    + \ps{V V}{6}T^{}_{Ijklmn}\etadnv_j\etadnv_k\eta_l\xi_m\xi_n
    \nonumber
    \\
    &
    + \ps{x V}{5}T^{}_{Ijklm}\etadnv_j\etadnv_k\xi_l\xi_m
    &&
    + \ps{U V}{6}T^{}_{Ijklmn}\etadnv_j\xi_k\etadnv_l\xi_m\xi_n
    \nonumber
    \\
    &
    + \ps{V V}{6}T^{}_{Ijklmn}\etadnv_j\eta_k \etadnv_l\xi_m\xi_n,
\end{alignat*}
where
{\color{black}
\begin{subequations}
\begin{align*}
    \ps{x x}{2}T^{}_{IJ} 
    & =
    4z_0^2 
    V_{iI} T_{ijkl} V_{jJ} x_{0,k}x_{0,l}
    \\
    \ps{x U}{3}T^{}_{IJK} 
    & =
    4z_0^2
    V_{iI} T_{ijkl} V_{jJ} x_{0,k} U_{lL}
    \\
    \ps{x V}{3}T^{}_{IJK} 
    & =
    4z_0^2
    V_{iI} T_{ijkl} V_{jJ} x_{0,k} V_{lL}
    \\
    \ps{U x}{3}T^{}_{IJK} 
    & =
    4z_0^2
    V_{iI} T_{ijkl} V_{jJ}  U_{kK} x_{0,l}
    \\
    \ps{U U}{4}T^{}_{IJKL} 
    & =
    4z_0^2
    V_{iI} T_{ijkl} V_{jJ}  U_{kK}  U_{lL}
    \\
    \ps{U V}{4}T^{}_{IJKL} 
    & =
    4z_0^2
    V_{iI} T_{ijkl} V_{jJ} U_{kK} V_{lL}
    \\
    \ps{V x}{3}T^{}_{IJK} 
    & =
    4z_0^2
    V_{iI} T_{ijkl} V_{jJ} V_{kK} x_{0,l}
    \\
    \ps{V U}{4}T^{}_{IJKL} 
    & =
    4z_0^2
    V_{iI} T_{ijkl} V_{jJ}  V_{kK}  U_{lL}
    \\
    \ps{V V}{4}T^{}_{IJKL} 
    & =
    4z_0^2
   V_{iI} T_{ijkl} V_{jJ}  V_{kK} V_{lL},
\end{align*}
\end{subequations}
}
{\color{black}
\begin{subequations}
\begin{align*}
    \ps{x x}{3}T^{}_{IJK} 
    & =
    8z_0 U^z_K
    V_{iI} T_{ijkl} V_{jJ} x_{0,k}x_{0,l} 
    \\
    \ps{x U}{4}T^{}_{IJKL} 
    & =
    8z_0 U^z_L
    V_{iI} T_{ijkl} V_{jJ} x_{0,k} U_{lL}
    \\
    \ps{x V}{4}T^{}_{IJKL} 
    & =
    8z_0 U^z_L
    V_{iI} T_{ijkl} V_{jJ} x_{0,k} V_{lL}
    \\
    \ps{U x}{4}T^{}_{IJKL} 
    & =
    8z_0 U^z_L
    V_{iI} T_{ijkl} V_{jJ}  U_{kK} x_{0,l}
    \\
    \ps{U U}{5}T^{}_{IJKLM} 
    & =
    8z_0 U^z_M
    V_{iI} T_{ijkl} V_{jJ}  U_{kK}  U_{lL}
    \\
    \ps{U V}{5}T^{}_{IJKLM} 
    & =
    8z_0 U^z_M
    V_{iI} T_{ijkl} V_{jJ} U_{kK} V_{lL}
    \\
    \ps{V x}{4}T^{}_{IJKL} 
    & =
    8z_0 U^z_L
    V_{iI} T_{ijkl} V_{jJ} V_{kK} x_{0,l}
    \\
    \ps{V U}{5}T^{}_{IJKLM} 
    & =
    8z_0 U^z_M
    V_{iI} T_{ijkl} V_{jJ}  V_{kK}  U_{lL}
    \\
    \ps{V V}{5}T^{}_{IJKLM} 
    & =
    8z_0 U^z_M
   V_{iI} T_{ijkl} V_{jJ}  V_{kK} V_{lL},
\end{align*}
\end{subequations}
}
\begin{subequations}
\begin{align*}
    \ps{x x}{4}T^{}_{IJKL} 
    & =
    4U^z_K U^z_L
    V_{iI} T_{ijkl} V_{jJ} x_{0,k}x_{0,l} 
    \\
    \ps{x U}{5}T^{}_{IJKLM} 
    & =
    4U^z_L U^z_M
    V_{iI} T_{ijkl} V_{jJ} x_{0,k} U_{lL}
    \\
    \ps{x V}{5}T^{}_{IJKLM} 
    & =
    4U^z_L U^z_M
    V_{iI} T_{ijkl} V_{jJ} x_{0,k} V_{lL}
    \\
    \ps{U x}{5}T^{}_{IJKLM} 
    & =
    4U^z_L U^z_M
    V_{iI} T_{ijkl} V_{jJ}  U_{kK} x_{0,l}
    \\
    \ps{U U}{6}T^{}_{IJKLMN} 
    & =
    4U^z_M U^z_N
    V_{iI} T_{ijkl} V_{jJ}  U_{kK}  U_{lL}
    \\
    \ps{U V}{6}T^{}_{IJKLMN} 
    & =
    4U^z_M U^z_N
    V_{iI} T_{ijkl} V_{jJ} U_{kK} V_{lL}
    \\
    \ps{V x}{5}T^{}_{IJKLM} 
    & =
    4U^z_L U^z_M
    V_{iI} T_{ijkl} V_{jJ} V_{kK} x_{0,l}
    \\
    \ps{V U}{6}T^{}_{IJKLMN} 
    & =
    4U^z_M U^z_N
    V_{iI} T_{ijkl} V_{jJ}  V_{kK}  U_{lL}
    \\
    \ps{V V}{6}T^{}_{IJKLMN} 
    & =
    4U^z_M U^z_N
    V_{iI} T_{ijkl} V_{jJ}  V_{kK} V_{lL},
\end{align*}
\end{subequations}
and
\begin{equation*}
    T_{IJKL} = - \frac{\pi}{4}  \rho w A_{mJ}R_{mp}B_{pK}R_{Is}B_{sL}.
\end{equation*}
\section{Additional simulation results}\label{sec:appendix_additional_optimization_results}

\MD{This section provides robustness checks regarding the simulation results presented in~ \Cref{sec:res_sim_setup}, grouped in three analysis. Firstly, \Cref{sec:appendix_error_profiles} extends the comparison between the FOM and the (P)ROMs for an alternative muscle placement. Secondly, \Cref{sec:appendix_rob_analysis} compares the ROB used in \Cref{sec:res_ROB_selection} to two other alternatives. Thirdly, the convergence of the solver and its solutions is analyzed in \Cref{sec:appendix_simulation_convergence}. }

\subsection{Model comparisons: error profiles}\label{sec:appendix_error_profiles}

\begin{figure}[htb]
    \centering
    \includegraphics[width=\linewidth]{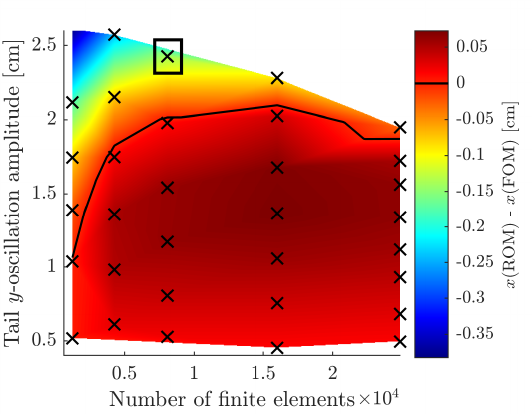}
    \caption{Absolute error in the swimming distance after $t=2$\si{s} between the FOM and the ROM for multiple discretizations and oscillation's amplitudes of the nominal shape (see \Cref{fig:A_muscles_placement_VM}). The crosses indicate simulations results, and the heat map is obtained using a linear interpolation between these results. Simulations of the FOM performed in the white area in the top right corner (i.e., for large number of FEs and large oscillations) did not converge. The rectangle corresponds to the relative error measured in \Cref{fig:A_accuracy}.}
    \label{fig:FOM_PROM_abs_error_profile}
\end{figure}

\MD{When comparing the simulation results from the FOM to the ones of the ROM, \Cref{fig:A_FOM_ROM_comparison} reports relative errors between -10 and 15\%. While the relative error grows for smaller tail oscillation, the absolute error remains small (\Cref{fig:FOM_PROM_abs_error_profile}).}

\begin{figure}[htb]
    \centering
    \includegraphics[width=\linewidth]{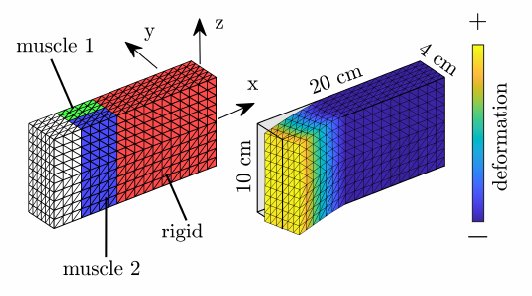}
    \caption{Alternative muscle placements (left) and corresponding vibration mode (right) used to obtain the error profile~\Cref{fig:FOM_PROM_alternative_error_profile}.}
    \label{fig:alternative_muscle_placements}
\end{figure}

\begin{figure}[htb]
    \centering
    \includegraphics[width=\linewidth]{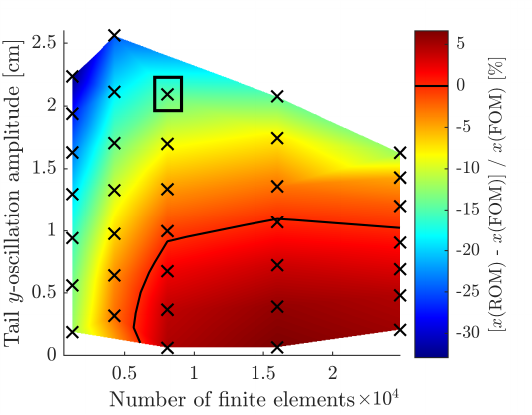}
    \caption{Relative error in the swimming distance after $t=2$\si{s} between the FOM and the ROM for multiple discretizations and oscillation's amplitudes of the nominal shape with alternative muscle placement (see \Cref{fig:alternative_muscle_placements}).}
    \label{fig:FOM_PROM_alternative_error_profile}
\end{figure}

\MD{To understand if these results are dependent on the muscle placement along the fish body, we consider the alternative arrangement shown in \Cref{fig:alternative_muscle_placements}. The relative errors between the FOM and ROM simulations have a similar magnitude than the results in \Cref{sec:res_model_comparisons}, with errors between -15\% and and 5\% for most combinations of oscillation and FE number (\Cref{fig:FOM_PROM_alternative_error_profile}. This increases the confidence in the reduced simulation results.}

\subsection{ROB selection analysis}\label{sec:appendix_rob_analysis}

\MD{As shown in \Cref{sec:res_model_comparisons}, and in particular in \Cref{fig:A_accuracy} and \Cref{fig:A_FOM_ROM_comparison}, using a (P)ROM instead of the FOM yields small errors regarding the tail oscillation and the swimming distance. In the following, we investigate how the ROB selection presented in \Cref{sec:res_ROB_selection} impacts these errors. We consider two variations of the ROB: (1) a ROB with the first VM but no MDs, and (2) a ROB with the first 2 VMs and their corresponding MDs. We select 6 combinations of mesh discretizations and actuation stiffnesses, and compare the results obtained using the FOM and the ROM. These combinations consists of the case marked with a rectangle in \Cref{fig:A_FOM_ROM_comparison}, as well as the cases around it (marked with crosses).}

\MD{
The results are presented in \Cref{tab:rel_error_VM1_wo_MDs} and \Cref{tab:rel_error_VM2_w_MDs}, where the third column is the reference ROB used in \Cref{sec:res_ROB_selection}, and the fourth column the alternative selection. Compared to the ROB used in \Cref{sec:06_results}, the ROB (1) without MDs shows poorer performance. The MDs allow to capture part of the nonlinearities of the FOM, which is not possible when using the first VM only. The ROB (2) with 2 VMS and their corresponding MDS is only marginally different than the ROB used in \Cref{sec:06_results}. This can easily be explained by the fact that the second VM, shown in \Cref{fig:second_VM}, is perpendicular to the deformation of the fish, and is in fact not activated when the fish is swimming. 
\begin{table}[ht]
\footnotesize
\caption{Relative error comparison between the FOM and two ROMs based on different ROBs: a ROB with one VM and its corresponding MDs (reference, \Cref{sec:res_ROB_selection}), and a ROB with one VM but no MDs. \label{tab:rel_error_VM1_wo_MDs}}
    \centering
    \renewcommand{\arraystretch}{1.3}
\rowcolors{3}{white}{gray!25}
    \begin{tabular}{cccc}
        \toprule
        \# FEs  & Tail    & Rel.\% error  & Rel.\% error \\
                & oscillation  [cm]     & w/ MDs        &  w/o  MDs\\
        \midrule
        4270  & 2.1533 & -3.9528 & -42.2850 \\
        4270  & 1.7495 &  1.2814 & -46.0000 \\
        8086  & 2.4272 & -5.9964 & -45.2700 \\
        8086  & 1.9787 &  0.7696 & -50.0070 \\
        16009 & 2.0252 &  2.1396 & -50.0420 \\
        16009 & 1.6775 &  6.0262 & -39.6960 \\
        \bottomrule
    \end{tabular}  
\end{table}
\begin{table}[ht]
\footnotesize
\caption{Relative error comparison between the FOM and two ROMs based on different ROBs: a ROB with one VM and its corresponding MDs (reference, \Cref{sec:res_ROB_selection}), and a ROB with two VMs and MDs.\label{tab:rel_error_VM2_w_MDs}}
\renewcommand{\arraystretch}{1.3}
\rowcolors{3}{white}{gray!25}
    \centering
    \begin{tabular}{cccc}
        \toprule
        \# FEs  & Tail    & Rel.\% error  & Rel.\% error \\
                & oscillation  [cm]     & 1 VM  &  2VMs \\
        \midrule
        4270  & 2.1533 & -3.9528 & -3.5768 \\
        4270  & 1.7495 &  1.2814 &  1.5406 \\
        8086  & 2.4272 & -5.9964 & -5.4423 \\
        8086  & 1.9787 &  0.7696 &  1.1668 \\
        16009 & 2.0252 &  2.1396 &  2.5219 \\
        16009 & 1.6775 &  6.0262 &  6.3103 \\
        \bottomrule
    \end{tabular}
\end{table}
\begin{figure}
    \centering
    \includegraphics[width=\linewidth, trim=0cm 0.9cm 0cm 0.1cm, clip]{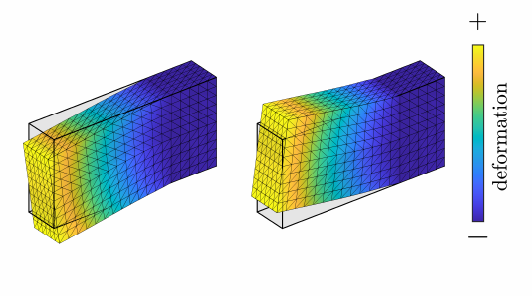}
    \caption{Second vibration mode, with a deformation perpendicular to the oscillation of the tail.}
    \label{fig:second_VM}
\end{figure}
}

\MD{
While adding VMs are not improving the accuracy of the ROM in this particular case, the choice of the ROB ultimately depends on the robotic system being modeled. More complex systems, such as the robotic arms used by \cite{katzschmann2019dynamically} for example, would require a more expressive ROBs with more VMs and MDs.
}
\subsection{Simulation convergence}\label{sec:appendix_simulation_convergence}

\MD{This section presents an analysis of the time stepping scheme followed in the simulations, assessing the convergence of the integration scheme and the validity of the used time step of 0.02\si{s}.}

\begin{figure}[htb]
    \centering
    \includegraphics[width=1\linewidth]{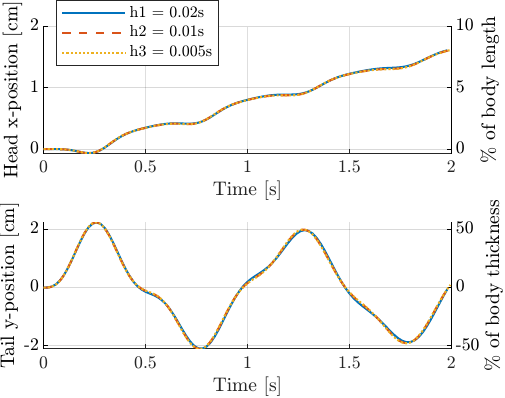}
    \caption{Trajectories of the nominal shape for different integration time steps, obtained using the PROM with 5 parameters.}
    \label{fig:convergent_trajectories}
\end{figure}

\MD{Considering the nominal shape discuss in \Cref{sec:res_sim_setup}, we solve the EoMs and obtain the sensitivies for the PROM with 5 parameters for three different integration time steps: $h=0.02$\si{s}, $h=0.01$\si{s}, $h=0.005$\si{s}. Firstly, \Cref{fig:convergent_trajectories} shows that the refinement of the time step does not change the physical behavior of the system.}

\MD{Secondly, we compute the following errors evaluated at common time instants
\begin{align*}
    e_{0.02\rightarrow0.01}(t) &= u_{0.02}(t)-u_{0.01}(t),\\ 
    e_{0.01\rightarrow0.005}(t) &= u_{0.01}(t)-u_{0.005}(t),
\end{align*}
where $u$ is the head $x$-position for the different integration time steps ($h=0.02$\si{s}, $h=0.01$\si{s}, $h=0.005$\si{s}). Taking the L2 norm over the vectors of errors gives the refinement ratio
\begin{equation*}
    \rho 
    = \frac{E_{0.02\rightarrow0.01}}{E_{0.01\rightarrow0.005}}
    =\frac{||e_{0.02\rightarrow0.01}||_2}{||e_{0.01\rightarrow0.005}||_2}=3.6,
\end{equation*}
which is consistent with second-order convergence, as expected using the Newmark-$\beta$ integration scheme. The solver therefore operates in asymptotic regime.}

\MD{Finally, we follow a similar procedure to show the convergence of the parameter sensitivities~\eqref{eq:def_sensitivity}. Sensitivities being matrices, we compute the Frobenius norm at each time step:
\begin{align*}
    e_{0.02\rightarrow0.01}(t) &= ||S_{0.02}(t)-S_{0.01}(t)||_\text{F},\\ 
    e_{0.01\rightarrow0.005}(t) &= ||S_{0.01}(t)-S_{0.005}(t)||_\text{F}.
\end{align*}
The resulting refinement ratio is
\begin{equation*}
    \rho  
    = \frac{E_{0.02\rightarrow0.01}}{E_{0.01\rightarrow0.005}}
    = \frac{||e_{0.02\rightarrow0.01}||_2}{||e_{0.01\rightarrow0.005}||_2}=2.7,
\end{equation*}
which also demonstrates monotonic convergence of the sensitivities under time step refinement.
}
\MD{Together, these results confirms correct solver behavior and justify the time step of $h=0.02$~\si{s}.}

\section{Additional optimization results}
\MD{This section presents additional optimization results that zoom into the parameters' initialization (\Cref{sec:appendix_opt_initialization}), the quality of the PROM approximations and the related gradiets (\Cref{sec:appendix_PROM_approx}), and the extension of the shape optimization to a larger design space (\Cref{sec:appendix_alternative_shapes})}

\subsection{Optimization initialization}\label{sec:appendix_opt_initialization}
\begin{figure}[htb]
    \centering
    \includegraphics[width=\linewidth]{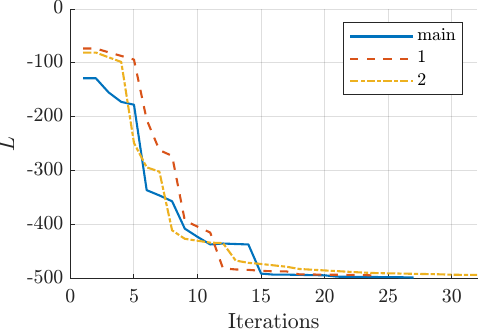}
    \caption{Cost function for SO1 with different random initializations of the search parameters (see \Cref{tab:optimization_initialization_SO1}).}
    \label{fig:param_init_cost_SO1}
\end{figure}

\begin{figure}[htb]
    \centering
    \includegraphics[width=\linewidth]{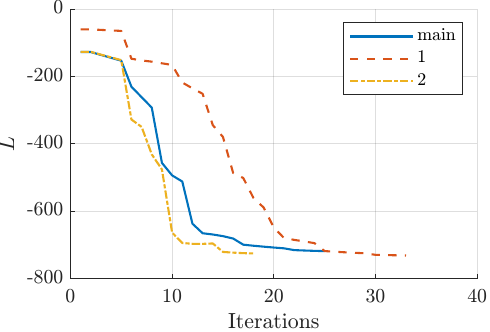}
    \caption{Cost function for SO2 with different random initializations of the search parameters (see \Cref{tab:optimization_initialization_SO2}).}
    \label{fig:param_init_cost_SO2}
\end{figure}

\begin{figure}[htb]
    \centering
    \includegraphics[width=\linewidth]{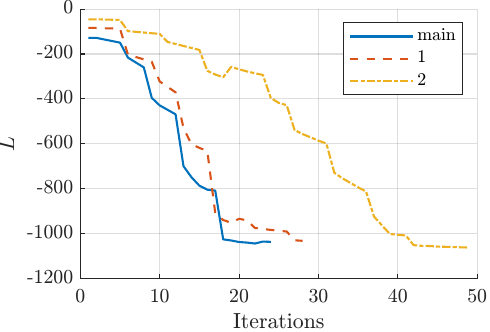}
    \caption{Cost function for SO3 with different random initializations of the search parameters (see \Cref{tab:optimization_initialization_SO3}).}
    \label{fig:param_init_cost_SO3}
\end{figure}

\MD{To assess how the shape parameters' initialization affects the optimization results, we run SO1-SO3 using two random set of initial parameters. The obtained optimal shapes are similar to the results obtained in \Cref{sec:shape_optimization}, as reported in \Cref{tab:optimization_initialization_SO1} (SO1), \Cref{tab:optimization_initialization_SO2} (SO2), and \Cref{tab:optimization_initialization_SO3} (SO3). The final costs are also identical, but the optimization trajectories might differ based on the parameter initialization (Figures \ref{fig:param_init_cost_SO1} to \ref{fig:param_init_cost_SO3}).}
\begin{table}[htb]
\footnotesize
\caption{Shape optimization results obtained with different initializations of the parameters for SO1. The columns denoted `main' contain the results presented in \Cref{sec:shape_optimization} and the columns denoted `1' and `2' contain results for two other parameters' initializations.\label{tab:optimization_initialization_SO1}
}
\begin{center}
\renewcommand{\arraystretch}{1.3}
\rowcolors{3}{white}{gray!25}
\begin{tabular}{c|ccc|ccc}
\toprule
 &  \multicolumn{3}{c|}{Initialization } & \multicolumn{3}{c}{Optimized parameters}  \\ 
        & main  & 1         & 2         & main      & 1         & 2         \\\midrule 
$\xi_1$ & $0$   & $-0.01$   & $-0.16$   & $-0.5$    & $-0.5$    & $-0.5$    \\
$\xi_2$ & $0$   & $-0.06$   & $-0.37$   & $0.29$    & $0.28$    & $0.27$    \\
$\xi_3$ & $0$   & $-0.13$   & $-0.09$   & $0.15$    & $0.15$    & $0.15$    \\\bottomrule
\end{tabular}
\end{center}
\end{table}

\begin{table}[htb]
\footnotesize
\caption{Shape optimization results obtained with different initializations of the parameters for SO2. The columns denoted `main' contain the results presented in \Cref{sec:shape_optimization} and the columns denoted `1' and `2' contain results for two other parameters' initializations.\label{tab:optimization_initialization_SO2}
}
\begin{center}
\renewcommand{\arraystretch}{1.3}
\rowcolors{3}{white}{gray!25}
\begin{tabular}{c|ccc|ccc}
\toprule
 &  \multicolumn{3}{c|}{Initialization } & \multicolumn{3}{c}{Optimized parameters}  \\ 
        & main  & 1         & 2         & main      & 1         & 2         \\\midrule 
$\xi_1$ & $0$   & $-0.01$   & $-0.16$   & $-0.49$   & $-0.50$   & $-0.50$   \\
$\xi_2$ & $0$   & $-0.06$   & $-0.37$   & $0.32$    & $0.31$    & $-0.44$   \\
$\xi_4$ & $0$   & $-0.13$   & $-0.09$   & $0.39$    & $0.40$    & $0.40$    \\
$\xi_5$ & $0$   & $-0.06$   & $-0.37$   & $0.49$    & $0.48$    & $0.49$    \\
$\xi_6$ & $0$   & $-0.13$   & $-0.09$   & $0.49$    & $0.49$    & $0.49$    \\\bottomrule
\end{tabular}
\end{center}
\end{table}

\begin{table}[htb]
\footnotesize
\caption{Shape optimization results obtained with different initializations of the parameters for SO3. The columns denoted `main' contain the results presented in \Cref{sec:shape_optimization} and the columns denoted `1' and `2' contain results for two other parameters' initializations.\label{tab:optimization_initialization_SO3}
}
\begin{center}
\renewcommand{\arraystretch}{1.3}
\rowcolors{3}{white}{gray!25}
\begin{tabular}{c|ccc|ccc}
\toprule
 &  \multicolumn{3}{c|}{Initialization } & \multicolumn{3}{c}{Optimized parameters}  \\ 
        & main  & 1         & 2         & main      & 1         &2          \\\midrule 
$\xi_1$ & $0$   & $-0.01$   & $-0.16$   & $-0.50$   & $-0.50$   & $-0.50$   \\
$\xi_2$ & $0$   & $-0.06$   & $-0.37$   & $0.32$    & $0.37$    & $0.42$    \\
$\xi_4$ & $0$   & $-0.13$   & $-0.09$   & $0.30$    & $0.29$    & $0.30$    \\
$\xi_5$ & $0$   & $-0.06$   & $-0.37$   & $0.39$    & $0.39$    & $0.38$    \\
$\xi_6$ & $0$   & $-0.13$   & $-0.09$   & $0.38$    & $0.37$    & $0.39$    \\
$\xi_7$ & $0$   & $-0.06$   & $-0.37$   & $0.24$    & $0.25$    & $0.25$    \\
$\xi_8$ & $0$   & $-0.13$   & $-0.09$   & $0.245$   & $0.25$    & $0.25$    \\
$\xi_9$ & $0$   & $-0.06$   & $-0.37$   & $0.17$    & $0.18$    & $0.18$    \\\bottomrule
\end{tabular}
\end{center}
\end{table}

\subsection{Quality of PROM approximations}\label{sec:appendix_PROM_approx}

\MD{The optimization pipeline \Cref{fig:optimization_pipeline} uses the PROM to compute gradients and approximate the solution of the EoMs for shape morphologies described by parameters that are close to the one of the nominal shape. If the current shape-varied mesh is too different from the nominal shape, the approximation of the system's dynamics using the sensitivities~\eqref{eq:approximation_using_sensitivity} might not be accurate.}

\MD{In the optimization pipeline, the approximation of the system's motion is important to obtain the gradient with which we update the parameters. We are therefore not interested in obtaining a good approximation of the EoMs solutions \emph{per se}, but rather by the gradient computed based on this approximation. If this quantity departs too drastically from the value that a local PROM build for the current mesh would provide, we would like to rebuild a PROM.}

\MD{The current implementation let the user define two thresholds that control when a PROM is rebuilt. First, if any of the shape parameter deviates more than \texttt{rebuildThreshold} from the parameter used for the last EoMs solve, a new PROM is constructed and the EoMs are re-solved. Second, a rebuild is triggered once a maximum number of iterations, specified by \texttt{nRebuild}, is reached. These parameters define a trade-off between computational effort and model fidelity: larger thresholds reduce the number of PROM reconstructions and EoMs solves, improving efficiency, whereas smaller thresholds maintain closer agreement between the reduced model and the current shape parameters, yielding more accurate state predictions and gradients. The choice of these parameters ultimately depends on the complexity of the optimization landscape.}

\begin{figure}[htb]
    \centering
    \includegraphics[width=\linewidth]{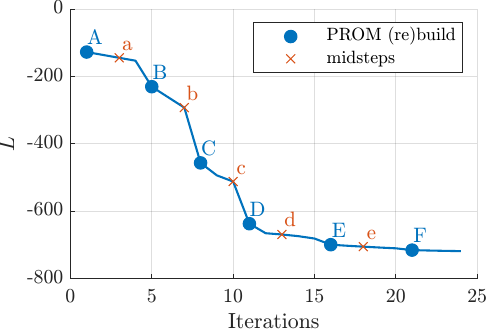}
    \caption{Cost function of SO1. Iterations at which a PROM is built are marked with circle and capital letters. Crosses indicates selected iterations at which approximation using PROM's sensitivities are used.}
    \label{fig:PROM_local_validity_cost_function_with_markers}
\end{figure}

\MD{The shape optimization experiments presented in \Cref{sec:shape_optimization} use \texttt{rebuildThreshold=0.15} and \texttt{nRebuild=5} or \texttt{nRebuild=6}. To assess the quality of the approximations and gradients obtained by the PROM under these specifications, we consider the case SO2, which cost function evolution is shown in \Cref{fig:PROM_local_validity_cost_function_with_markers}. Uppercase letters denote iteration at which a new PROM has been built. Lowercase letters represent selected midstep iterations at which approximations have been used. For each segment `IiJ', we compare two approximations obtained using the sensitivities~\eqref{eq:approximation_using_sensitivity} to ground-truth results obtained with local PROMs.}

\begin{figure*}[tb]
    \centering
    \begin{subfigure}{0.47\textwidth}
        \centering
        \includegraphics[width=\linewidth]{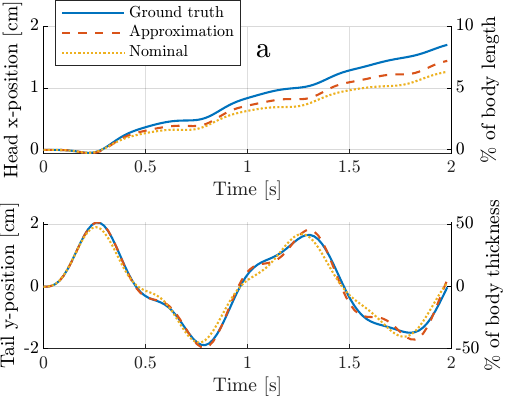}
    \end{subfigure}
    \hfill
    \begin{subfigure}{0.47\textwidth}
        \centering
        \includegraphics[width=\linewidth]{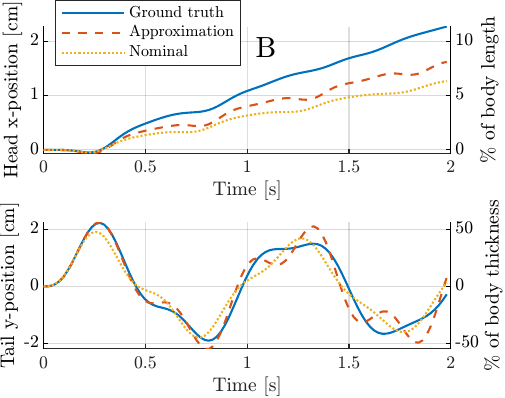}
    \end{subfigure}    
    \begin{subfigure}{0.47\textwidth}
        \centering
        \includegraphics[width=\linewidth]{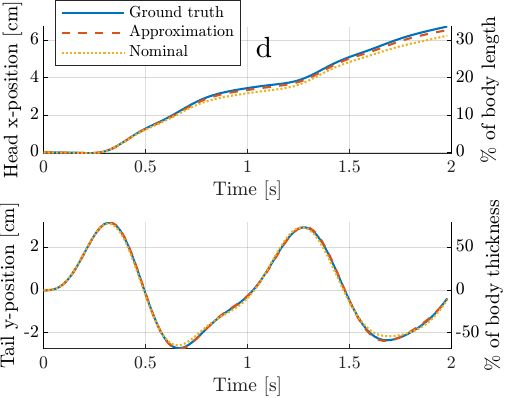}
    \end{subfigure}
    \hfill
    \begin{subfigure}{0.47\textwidth}
        \centering
        \includegraphics[width=\linewidth]{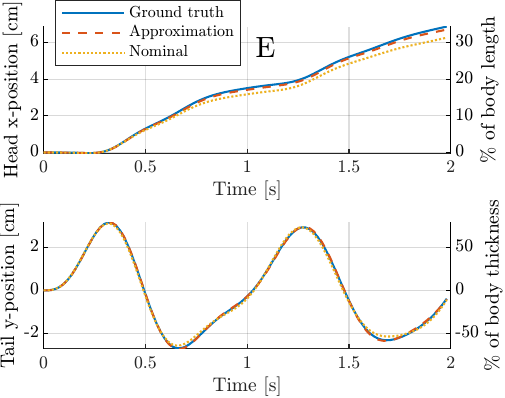}
    \end{subfigure}
    \caption{Comparison between ground truth motion provided by a local PROM and its approximation using sensitivities from a nominal PROM for selected iterations of SO2 (see \Cref{fig:PROM_local_validity_cost_function_with_markers}). \label{fig:PROM_approx_selected_cases}}
\end{figure*}

\MD{The PROM approximations are characterized by three features according to our results. First, their quality is poorer for segments `IjJ' at which a large cost decrease occurs (i.e., AaB, BbC, and CcD), and better close to convergence (DdE, EeF). \Cref{fig:PROM_approx_selected_cases} illustrates this pattern. Second, the error from the approximation is smaller at midsteps, and larger at rebuild step, when a new PROM is in fact rebuild (\Cref{fig:PROM_approx_selected_cases}, top panels, and \Cref{tab:PROM_gradient_comparison}. Third, the gradient obtained using the approximation is consistent in its direction with the ground truth gradient obtained by a local PROM, indicating that we can rely on the descent direction (\Cref{tab:PROM_gradient_comparison}).}

\begin{table}[ht]
\footnotesize
\caption{Comparison between the gradients obtained with PROM approximations ($\Tilde{\nabla L}$) and  ground truth local PROMs ($\nabla L$) for selected iterations of SO2 (see \Cref{fig:PROM_local_validity_cost_function_with_markers}). \label{tab:PROM_gradient_comparison}}
    \centering
    \renewcommand{\arraystretch}{1.3}
\rowcolors{2}{white}{gray!25}
    \begin{tabular}{c|cccc}
        \toprule
        It.  & $\nabla L$    & $\Tilde{\nabla L}$ & Rel. error [\%] & $\Delta$ angle [°] \\
        \midrule
        a  & $\left[\begin{array}{r}137 \\ -16 \\ -121 \\ -16 \\ -81\end{array}\right]$ & $\left[\begin{array}{r} 112 \\ -11 \\ -76 \\ -11 \\ -54 \end{array}\right]$  
        & $\left[\begin{array}{r} -18 \\ 30 \\ 37 \\ 32 \\ 34 \end{array}\right]$ & 7.4\\
        B & $\left[\begin{array}{r} 168 \\ -23 \\ -186 \\ -24 \\ -118 \end{array}\right]$
         & $\left[\begin{array}{r} 112 \\ -11 \\ -76 \\ -11 \\ -54 \end{array}\right]$&
        $\left[\begin{array}{r} -33 \\ 51 \\ 59 \\ 54 \\ 55 \end{array}\right]$
        & 13.2 \\
        d  & $\left[\begin{array}{r} 324 \\ -53 \\ -747 \\ -88 \\ -198 \end{array}\right]$
     & $\left[\begin{array}{r} 295 \\ -66 \\ -692 \\ -92 \\ -243 \end{array}\right]$
     & $\left[\begin{array}{r} -9 \\ -25 \\ 7 \\ -5 \\ -22 \end{array}\right]$
     & 4.4\\
            E  & $\left[\begin{array}{r} 323 \\ -30 \\ -774 \\ -68 \\ -209 \end{array}\right]$
    &  $\left[\begin{array}{r} 295 \\ -66 \\ -692 \\ -92 \\ -243 \end{array}\right]$
     & $\left[\begin{array}{r} -9 \\ -118 \\ 11 \\ -37 \\ -16 \end{array}\right]$
 & 5. 28\\
        \bottomrule
    \end{tabular}  
\end{table}

\subsection{Alternative shape variations}\label{sec:appendix_alternative_shapes}

\begin{figure*}
    \centering
    \includegraphics[width=\linewidth]{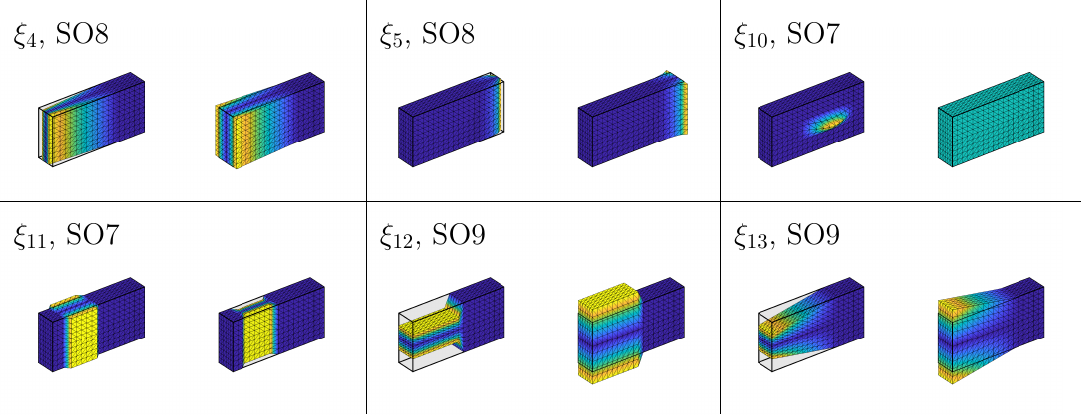}
    \caption{Additional shape variations used in the numerical experiments described in \Cref{sec:appendix_alternative_shapes}.}
    \label{fig:additional_shape_variations}
\end{figure*}

\MD{We present three additional case-studies focusing on shape optimization (SO7, SO8, and SO9). These cases use some of the shape variations depicted in \Cref{fig:all_shape_variations} as well as additional shape variations proposed in \Cref{fig:additional_shape_variations}. The exact selection and constraints for each of the three cases is summarized in \Cref{tab:constraint_summary_appendix}, resulting in cases with 5 or 6 optimization parameters.}
\begin{table}[tb]
\footnotesize
\caption{Lower and upper bound on the search parameters for shape optimization experiments SO7-SO9.\label{tab:constraint_summary_appendix}
}
\begin{center}
\renewcommand{\arraystretch}{1.3}
\rowcolors{2}{white}{gray!25}
\begin{tabular}{c|ccc}
\toprule
Case &  SO7 & SO8 & SO9 \\ \midrule 
$\xi_1$ & $-0.5, 0.5$     & $-0.5, 0.5$   &  $-0.5, 0.5$  \\
$\xi_2$ & $-0.5, 0.5$     & $-0.5, 0.5$   &  $-0.5, 0.5$  \\
$\xi_{3}$ & $-0.15, 0.15$ & $-0.15, 0.15$ &  not used  \\
$\xi_{4}$ & not used      & $-0.4, 0.4$   &  $-0.4, 0.4$  \\
$\xi_{5}$ & not used      & $-0.4, 0.4$   &  not used \\
$\xi_{6}$ & not used      & not used      &  $-0.4, 0.4$ \\
$\xi_{10}$ & $-0.02, 0.3$  & not used      &  not used  \\
$\xi_{11}$ & $-0.3, 0.3$  & not used      &  not used  \\
$\xi_{12}$ & not used     & not used      &  $-0.5, 0.5$  \\
$\xi_{13}$ & not used     & not used      &  $-0.5, 0.5$  \\\bottomrule
\end{tabular}
\end{center}
\end{table}

\begin{figure}[htb]
    \centering
    \includegraphics[width=\linewidth]{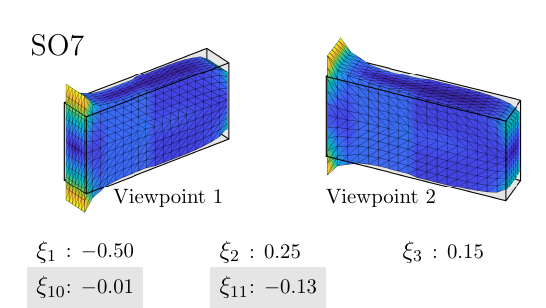}
    \caption{Optimal shape obtained for SO7.}
    \label{fig:SO7}
\end{figure}

\begin{figure}[htb]
    \centering
    \includegraphics[width=\linewidth,trim={0 0.05cm 0 0.5cm},
        clip]{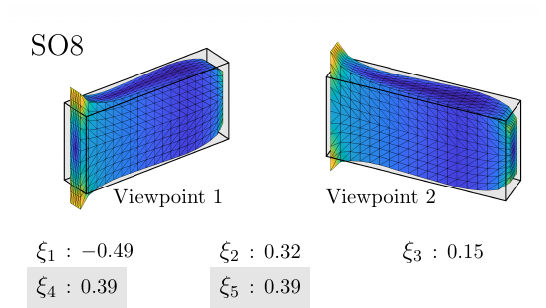}
    \caption{Optimal shape obtained for SO8.}
    \label{fig:SO8}
\end{figure}

\MD{First, we compare the results of SO7 and SO8, which both have 5 shape variations that represent lateral features of the fish. While SO7 comprises a lateral notch ($\xi_{11}$) and a fin ($\xi_{10}$), SO8 comprises the feature of a thiner lateral the tail ($\xi_4$) and rounded corners ($\xi_6$). While SO8 fully uses the search space and hits the constraints (\Cref{fig:SO8}, parmeters of interest highlighted in gray), SO7 does not heavily use the fin and the notch features (\Cref{fig:SO7}). The swimming distance is also considerably larger for SO8 compared to SO7 (9.1\si{cm} compared to 6.3\si{cm}). }

\MD{This example shows that not only the number of shape variations matters for the optimal swimming performance, but also their ability to express meaningful features. While the optimization pipeline can accomodate any type of shape variations, they should be chosen according with some knowledge of the test case to be expressive.}

\begin{figure}[htb]
    \centering
    \includegraphics[width=\linewidth,trim={0 0.05cm 0 0.5cm},
        clip]{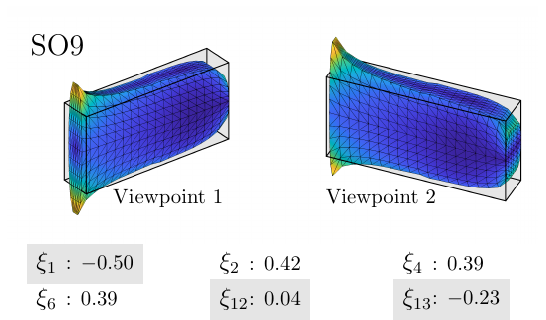}
    \caption{Optimal shape obtained for SO9.}
    \label{fig:SO9}
\end{figure}

\MD{The last example provided by SO9 focuses on the feature selection of a tail under active design constraint. Three possible tail features can be combined ($\xi_1, \xi_{12}, \xi_{13}$) but the height of the tail tip cannot exceed a certain threshold, leading to the joint constraint $-0.8\leq \xi_1 +\xi_{12} + \xi_{13}\leq 0.8$. The optimizer therefore has to weight competing designs for a limited amount of physical space. The resulting morphology is shown in \Cref{fig:SO9} and reached 7.3~\si{cm}. As shown in the gray cells of \Cref{fig:SO9}, the optimizer uses mainly the shape variation 1 and 13, but leaves the feature 12 aside. Such an example shows that the pipeline can be practically useful to decide between competing morphological features.}









\end{document}